\PassOptionsToPackage{table}{xcolor}
\PassOptionsToPackage{numbers,square,sort&compress}{natbib}
\pdfoutput=1
\documentclass{article}
\ifdefined\buildmode\else\def\buildmode{all}\fi   %
\def\buildmodemanuscript{manuscript}\def\buildmodesupplement{supplement}
\newif\ifmanuscriptpart\newif\ifsupplementpart
\ifx\buildmode\buildmodemanuscript \manuscriptparttrue\supplementpartfalse
\else\ifx\buildmode\buildmodesupplement \manuscriptpartfalse\supplementparttrue
\else \manuscriptparttrue\supplementparttrue \fi\fi

\usepackage[T1]{fontenc}
\usepackage[utf8]{inputenc}
\makeatletter

\RequirePackage{tgpagella} %
\RequirePackage{mathpazo}  %
\RequirePackage{inconsolata} %

\usepackage{eso-pic} %
\RequirePackage{fancyhdr}
\RequirePackage{xcolor}
\RequirePackage{natbib}
\RequirePackage[letterpaper,margin=2.5cm,headheight=24pt]{geometry}

\IfFileExists{datetime2.sty}{\RequirePackage[style=iso]{datetime2}}{}
\RequirePackage{tikz}
\RequirePackage{titling}

\RequirePackage{microtype}
\RequirePackage{setspace}
\RequirePackage{hyperref}
\RequirePackage{url}
\definecolor{darkblue}{rgb}{0, 0, 0.5}
\hypersetup{colorlinks=true, citecolor=darkblue, 
linkcolor=black, urlcolor=darkblue}

\setcitestyle{numbers,square,citesep={,}}

\def\maketitle{\par
\begingroup
   \def\thefootnote{\fnsymbol{footnote}}
   \def\@makefnmark{\hbox to 0pt{$^{\@thefnmark}$\hss}} %
   \long\def\@makefntext##1{\parindent 1em\noindent
                            \hbox to1.8em{\hss $\m@th ^{\@thefnmark}$}##1}
   \@maketitle \@thanks
\endgroup
\setcounter{footnote}{0}
\let\maketitle\relax \let\@maketitle\relax
\gdef\@thanks{}\gdef\@author{}\gdef\@title{}\let\thanks\relax}

\usepackage{fancyhdr}
\renewcommand{\headrulewidth}{1pt}
\renewcommand{\headrule}{{\color[HTML]{4D4D4D}\hrule\@height\headrulewidth\@width\headwidth\vskip-\headrulewidth}}

\def\twodigits#1{\ifnum#1<10 0\fi\the#1}
\fancypagestyle{firstpage}{
  \lhead{\includegraphics[height=20pt]{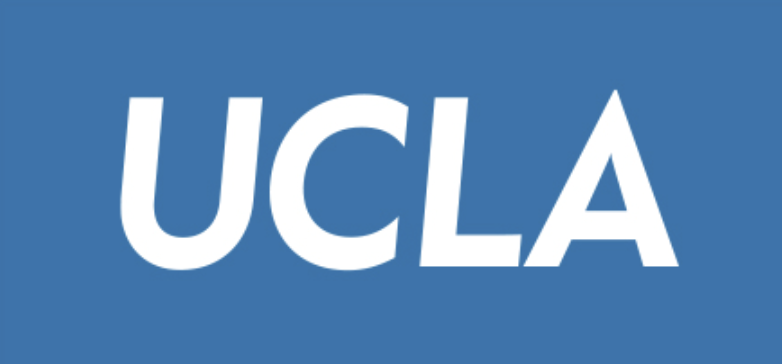}\hspace{0.5em}\includegraphics[height=20pt]{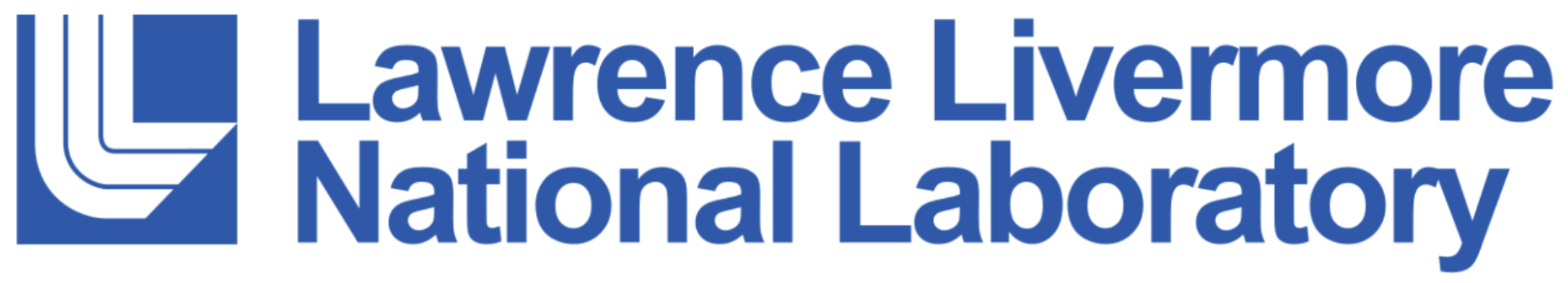}}
  \rhead{\textcolor[HTML]{4D4D4D}{\today}}
  \renewcommand{\headrule}{{\color[HTML]{4D4D4D}\hrule\@height\headrulewidth\@width\headwidth\vskip-\headrulewidth}}
  \lfoot{}
  \cfoot{\thepage}
  \rfoot{}
  
}

\def\@maketitle{\vbox{\hsize\textwidth
\vskip 0.1in
\begin{center}%
{\Large\bfseries\@title\par}%
\end{center}%
\vspace{-0.3cm}
 \begin{quote}\rule{\z@}{6pt} {\centering\par{\@author}\par}\end{quote}%
\vskip 0.3in minus 0.1in
}
\thispagestyle{firstpage}}

\renewenvironment{abstract}{\vskip.075in\centerline{\large\bf
Abstract}\vspace{0.5ex}\begin{quote}}{\par\end{quote}\vskip 1ex}

\renewenvironment{abstract}
  {\begin{center}
    \begin{tcolorbox}[
        colframe=abstractblue,      %
        colback=white,          %
        arc=2mm,                %
        boxrule=0.9pt,          %
        left=1mm, right=1mm,    %
        top=3mm, bottom=4mm,    %
        enhanced,
        width=0.95\textwidth    %
      ]
      \centering{\large\bf\color{abstractblue} Abstract}\par\vspace{1ex}
      \begin{quote}
  }
  {\end{quote}
    \end{tcolorbox}
  \end{center}
}

\definecolor{richpurple}{RGB}{75,46,131}
\definecolor{abstractblue}{RGB}{0,82,155}

\definecolor{mypurple}{RGB}{102,48,227}
\definecolor{deeppurple}{RGB}{115,51,161}

\def\section{\@startsection {section}{1}{\z@}{-2.0ex plus
    -0.5ex minus -.2ex}{1.5ex plus 0.3ex
minus0.2ex}{\large\bf\raggedright\color{black}}}

\def\subsection{\@startsection{subsection}{2}{\z@}{-1.8ex plus
-0.5ex minus -.2ex}{0.8ex plus .2ex}{\normalsize\bf\raggedright\color{black}}}
\def\subsubsection{\@startsection{subsubsection}{3}{\z@}{-1.5ex
plus      -0.5ex minus -.2ex}{0.5ex plus
.2ex}{\normalsize\bf\raggedright\color{black}}}
\def\paragraph{\@startsection{paragraph}{4}{\z@}{1.5ex plus
0.5ex minus .2ex}{-1em}{\normalsize\bf\color{black}}}
\def\subparagraph{\@startsection{subparagraph}{5}{\z@}{1.5ex plus
  0.5ex minus .2ex}{-1em}{\normalsize\color{black}}}

\skip\footins 9pt plus 4pt minus 2pt
\def\footnoterule{\kern-3pt \hrule width 12pc \kern 2.6pt }
\leftmargini\leftmargin \leftmarginii 2em
\def\@listi{\leftmargin\leftmargini}
\def\@listii{\leftmargin\leftmarginii
   \labelwidth\leftmarginii\advance\labelwidth-\labelsep
   \topsep 2pt plus 1pt minus 0.5pt
   \parsep 1pt plus 0.5pt minus 0.5pt
   \itemsep \parsep}
\def\@listiii{\leftmargin\leftmarginiii
    \labelwidth\leftmarginiii\advance\labelwidth-\labelsep
    \topsep 1pt plus 0.5pt minus 0.5pt
    \parsep \z@ \partopsep 0.5pt plus 0pt minus 0.5pt
    \itemsep \topsep}
\def\@listiv{\leftmargin\leftmarginiv
     \labelwidth\leftmarginiv\advance\labelwidth-\labelsep}
\def\@listv{\leftmargin\leftmarginv
     \labelwidth\leftmarginv\advance\labelwidth-\labelsep}
\def\@listvi{\leftmargin\leftmarginvi
     \labelwidth\leftmarginvi\advance\labelwidth-\labelsep}

\belowdisplayskip \abovedisplayskip
\def\normalsize{\@setsize\normalsize{11pt}\xpt\@xpt}
\def\small{\@setsize\small{10pt}\ixpt\@ixpt}
\def\footnotesize{\@setsize\footnotesize{10pt}\ixpt\@ixpt}
\def\xsmall{\@setsize\xsmall{8.8pt}\viipt\@viipt}
\def\scriptsize{\@setsize\scriptsize{8pt}\viipt\@viipt}
\def\tiny{\@setsize\tiny{7pt}\vipt\@vipt}
\def\large{\@setsize\large{14pt}\xiipt\@xiipt}
\def\Large{\@setsize\Large{16pt}\xivpt\@xivpt}
\def\LARGE{\@setsize\LARGE{20pt}\xviipt\@xviipt}
\def\huge{\@setsize\huge{23pt}\xxpt\@xxpt}
\def\Huge{\@setsize\Huge{28pt}\xxvpt\@xxvpt}

\makeatother
\usepackage[most]{tcolorbox}

\usepackage{amsmath,amssymb,amsthm}
\usepackage{graphicx}
\usepackage{booktabs,tabularx,array,longtable,multirow}

\usepackage{setspace}

\usepackage{caption}

\usepackage{enumitem}
\usepackage{algorithm,algpseudocode}
\usepackage{listings}
\usepackage{placeins}
\usepackage{xurl}
\usepackage{lineno}
\nolinenumbers

\DeclareCaptionLabelSeparator{pipe}{\ \textbar\ }
\floatname{algorithm}{Supplementary Algorithm}

\definecolor{docnotelinkcolor}{HTML}{1F3F7A}
\providecommand{\urlprefix}{}
\newcommand{\Ansatze}{Ans\"atze}
\newcommand{\code}[1]{\texttt{#1}}

\newcommand{\stab}[1]{\setcounter{table}{\numexpr#1-1\relax}}%
\newcolumntype{L}[1]{>{\raggedright\arraybackslash}p{#1}}
\newcolumntype{Y}{>{\raggedright\arraybackslash}X}
\newtheorem{proposition}{Proposition}

\newcommand{\seniorauthor}{\textsuperscript{\textdagger}}
\newcommand{\authoraffiliation}[1]{\textsuperscript{#1}}
\newcommand{\paperauthors}{%
  Eric~Hanchen~Jiang\authoraffiliation{1*\textdaggerdbl},
  Junyi~Zhang\authoraffiliation{1*\textdaggerdbl}, Jinxi~Yu\authoraffiliation{1*}, Jiachen~Lu\authoraffiliation{1*}, 
  Zhi~Zhang\authoraffiliation{1*},
  Michael~Simon\authoraffiliation{1},
  Justin~Baker\authoraffiliation{1},
  Richard~Cai\authoraffiliation{1},
  Sneha~Agarwal\authoraffiliation{1},
  Yubei~Li\authoraffiliation{1},
  Hyunsik~Chae\authoraffiliation{1},
  Xinjie~He\authoraffiliation{1},
  Kyle~Zheng\authoraffiliation{1},
  Vigyan~Sahai\authoraffiliation{1},
  Ethan~Ji\authoraffiliation{1}, Alexander~K~Taylor\authoraffiliation{1}, Yiwen~Kou\authoraffiliation{1},
  Kyle~Hess\authoraffiliation{1},
  Xiao~Liang\authoraffiliation{1},
  Yingnian~Wu\authoraffiliation{1},
  J.D.~Zamfirescu-Pereira\authoraffiliation{1},
  Andrea~Bertozzi\authoraffiliation{1}\seniorauthor,
  Kai-Wei~Chang\authoraffiliation{1}\seniorauthor,
  Eunice~Jun\authoraffiliation{1}\seniorauthor,
  Raghu~Meka\authoraffiliation{1}\seniorauthor,
  Nanyun~Peng\authoraffiliation{1}\seniorauthor,
  Amit~Sahai\authoraffiliation{1}\seniorauthor,
  Matthew~Sottile\authoraffiliation{2}\seniorauthor,
  Terence~Tao\authoraffiliation{1}\seniorauthor,
  Wei~Wang\authoraffiliation{1}\seniorauthor
  }
\newcommand{\paperaffiliations}{%
  \textsuperscript{1}University of California, Los Angeles (UCLA)\\[.35ex]
  \textsuperscript{2}Lawrence Livermore National Laboratory}

\setcitestyle{numbers,square,comma}
\hypersetup{colorlinks=true,citecolor=darkblue,linkcolor=darkblue,urlcolor=darkblue,
  pdftitle={Continual Graph Memory and Community Infrastructure for AI-Assisted Mathematical Discovery},pdfauthor={\paperauthors},
  pdfsubject={Persistent mathematical research agents and a community research platform},
  pdfkeywords={mathematical research, research agents, graph memory, human--AI collaboration}}
\newcommand{\revisiondate}{6 November 2026}
\renewcommand{\today}{\revisiondate}
\title{Continual Graph Memory and Community Infrastructure for AI-Assisted Mathematical Discovery}
\author{%
  \parbox{\linewidth}{\centering\fontsize{10.5}{13}\selectfont\paperauthors}\\[13pt]
  \parbox{\linewidth}{\centering\paperaffiliations\\[.65ex]
  {\small\href{https://ansatze.ai}{\raisebox{-.2em}{\includegraphics[height=1.05em]{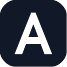}}\hspace{.4em}ansatze.ai}\quad\textbullet\quad\href{https://github.com/ericjiang18/agent_monitor_math}{\raisebox{-.2em}{\includegraphics[height=1.05em]{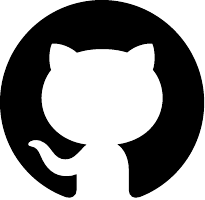}}\hspace{.4em}Platform Repo}\quad\textbullet\quad\href{https://github.com/uclanlp/continual-graph-memory}{\raisebox{-.2em}{\includegraphics[height=1.05em]{figures/link-icons/github.pdf}}\hspace{.4em}Harness Repo}}}}

\begin{document}
\ifmanuscriptpart
\maketitle
\begingroup
\renewcommand{\thefootnote}{\fnsymbol{footnote}}
\footnotetext[3]{Project leads: Eric Hanchen Jiang and Junyi Zhang.} \footnotetext[1]{Equal Contribution}\footnotetext[2]{Senior authors served as equal advisors and are listed alphabetically}
\endgroup
\begingroup
\renewcommand{\thefootnote}{}
\footnotetext[0]{For M. Sottile, this work was performed under the auspices of the U.S. Department of Energy by Lawrence Livermore National Laboratory under Contract DE-AC52-07NA27344}
\endgroup
\begin{abstract}
Mathematical discovery requires preserving partial ideas and the evidence supporting their reuse. We present Ansatz, a research harness whose continual graph memory separates model-accepted claims and dependencies from curated exploration, and \Ansatze, a community platform for inspecting, guiding and checking this work. On First Proof, Ansatz recorded nine local verifier closures in each of two ten-problem batches; a partial result in the Second Batch also counted as a pass under the study's criterion. These model-assessed outcomes include continued searches and differ from expert validation. Component studies and re-review expose disagreements and their dependency reach. Open-problem investigations produced candidate results, with five Lean formalizations reported. The platform connects problem discovery, research records, manuscript analysis and formal checking while preserving distinct evidence states. Together, these systems support cumulative human--AI research through persistent investigation and inspectable contributions.
\end{abstract}
\section{Introduction}

Mathematical research advances by connecting partial insights. A lemma from an abandoned approach may unlock a proof, a counterexample may expose a missing hypothesis, and a reduction may become useful much later. Artificial intelligence systems have discovered mathematical constructions~\cite{funsearch,georgiev2025}, achieved olympiad-level theorem proving~\cite{alphageometry,alphaproof} and pursued research problems through extended searches~\cite{danus,albilich,rethlas}. Sustaining these investigations requires a record of which claims were accepted, what they depend on, which approaches remain unresolved and why others failed. A final answer alone cannot preserve this research state.

The same problem becomes consequential when work enters a community. Mathematical contributions acquire value through scrutiny, attribution and reuse; collaborative investigations and formal libraries show how intermediate work can support subsequent progress~\cite{gowers2009,mathlib}. Researchers with different expertise can inspect a reduction, repair a gap or redirect an unsuccessful search. Yet sharing can also propagate an unsupported premise or detach a conclusion from its assumptions and checking history. A community platform therefore needs to preserve contributions together with the evidence required to challenge and revise them. Explicit dependencies and provenance identify the object of review~\cite{provdm,fair}, although neither establishes mathematical validity.

We introduce Ansatz, a persistent research harness built around continual graph memory, and \Ansatze, a community platform that makes the resulting research state inspectable~\cite{ansatzeai,platformrepo}. Ansatz connects a graph of model-accepted claims and declared dependencies with a separately curated exploration graph of plans, findings, obstacles, counterexamples and other intermediate results of exploration. Parallel workers propose arguments, a fresh verifier reviews submissions, a main agent coordinates work and a curator maintains the exploratory record. Admission, revocation and retrieval govern how contributions persist across sessions. Cross-project recall supplies leads that must be established again before they can serve as premises.

\Ansatze\ extends this research model to problem discovery, argument inspection, feedback, manuscript analysis and formal checking. It separates execution status, model review, dependency closure, formal verification, statement fidelity and provenance (Extended Data Table~\ref{tab:evidence}). This distinction matters because a completed run can leave its target unresolved, an accepted argument can rely on a disputed premise, and a checked formal proof can express a different statement from the intended problem. An accessible workspace must expose these distinctions if community members are to decide where their attention is useful.

We examine this framework on twenty First Proof benchmark problems~\cite{firstproof2,firstproof3}, component studies, open-problem investigations and platform implementation checks. The evidence characterizes persistent search and the records available for subsequent review. 

\begin{figure}[t]
\centering
\includegraphics[width=\linewidth,height=0.67\textheight,keepaspectratio]{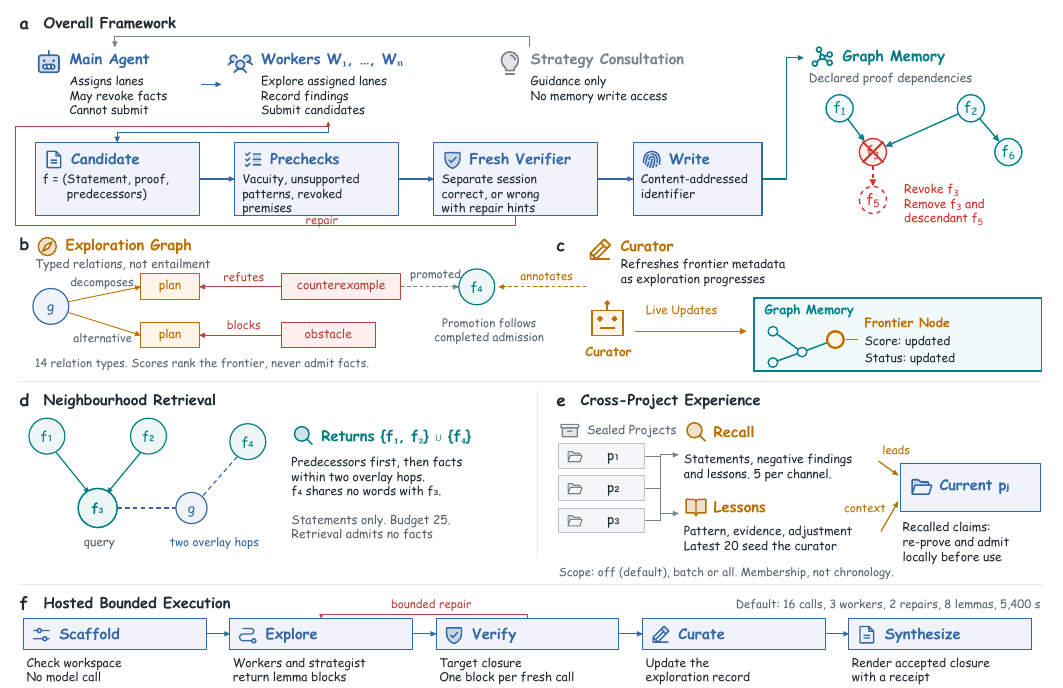}
\caption{\textbf{Ansatz framework and continual graph memory.}
\textbf{a},~Workers submit candidates for prechecks, independent verification and admission to graph memory; revocation removes declared descendants.
\textbf{b},~The exploration graph connects plans, findings, obstacles, counterexamples and other intermediate results of exploration through typed relations.
\textbf{c},~The curator continually updates frontier-node scores and status as exploration progresses.
\textbf{d},~Neighbourhood retrieval returns declared predecessors and nearby accepted facts.
\textbf{e},~Cross-project recall and lessons support subsequent exploration; recalled claims require local proof before use.
\textbf{f},~Hosted execution proceeds through scaffold, explore, verify, curate and synthesize stages under explicit budgets.}
\label{fig:cgm}
\end{figure}

\section{Methods}

\subsection{Harness framework and exploration graph}
Ansatz coordinates a main agent, parallel mathematical workers, a separate verifier and a resident curator around a persistent project workspace (Fig.~\ref{fig:cgm}a). The main agent assigns research directions and can obtain strategy consultation; workers develop arguments and publish intermediate results; the verifier reviews candidate facts; and the curator maintains the research frontier. Each worker retains private notes, while shared findings, accepted facts and graph annotations persist across sessions. A new worker session reads its current assignment and recovers relevant research state from these stores, allowing an investigation to continue beyond an individual context window.

The shared memory contains both a fact graph of accepted statements and their declared proof dependencies, and an exploration graph connecting plans, directions, conclusions, examples, counterexamples, proof attempts, obstacles, dead ends, strategic guidance, progress syntheses, verification records and reports, together with accepted facts and symbolic goals (Fig.~\ref{fig:cgm}b). Exploration nodes refer to existing shared findings, accepted facts or symbolic goals, identified by \code{goal:<slug>}. This reuses the underlying records rather than maintaining a separate copy of their mathematical content. A finding records its type, claim, evidence, author, time and optional glossary; objectively checkable findings require explicit evidence, whereas plans and directions record proposed lines of investigation. Relations and curator annotations are stored as separate append-only event logs, from which adjacency, current annotations and the frontier are reconstructed on demand.

Workers declare relations when publishing findings so that the graph retains how an investigation develops. Fourteen relation types encode complementary aspects of this process: \code{addresses}, \code{decomposes-into}, \code{alternative-of} and \code{depends-on} organize goals and approaches; \code{supports}, \code{refutes} and \code{blocks} attach evidence and obstacles; \code{refines} and \code{promoted-to} connect successive research stages; \code{supersedes} and \code{prioritizes} record changes in guidance; and \code{reduces-to}, \code{equivalent-to} and \code{generalizes} express transformations of the goal. Each edge retains its endpoints, type, author, timestamp and rationale. Thus, a failed route remains connected to the evidence that ruled it out, while a useful intermediate result remains associated with the goals it may support.

Decompositions carry an \code{and\_group} identifier that groups jointly required subgoals, with alternative plans represented as separate routes. An optional technique tag identifies obligations associated with localization, invariant arguments, extremal constructions, compactness or witness construction. For example, a localization plan includes both local statements and a reassembly step in its AND group. Reduction and reformulation edges record the identifiers of their supporting bridge or equivalence facts in the rationale. These relations preserve the obligations behind a proposed strategy; accepted proof dependencies remain recorded separately in the fact graph.

\subsection{Curator design and frontier maintenance}
The curator is a dedicated agent that maintains the evolving exploration state while workers continue their investigations (Fig.~\ref{fig:cgm}c). Its gateway permissions allow it to read shared findings and facts, inspect the frontier, append node annotations and record retrospective lessons. It cannot publish findings, declare relations, submit or revoke facts, or reassign workers through these tools. This division assigns exploration to workers, fact admission to the verification pathway, strategic coordination to the main agent and frontier assessment to the curator.

For each assessed node, the curator records a status, an optional score in $[0,1]$, the identifiers of evidence considered, an explanatory comment and a timestamp. Status takes one of five values: \code{open}, \code{active}, \code{blocked}, \code{dead} or \code{closed}. The score expresses the estimated promise of pursuing the node towards the project target. The operating contract requires the comment to explain this judgment using cited evidence, and instructs the curator to fetch complete findings and verification reports before assessing them. New annotations preserve the history of earlier assessments; the most recently appended event supplies the current view.

Updates proceed asynchronously in batches. The curator is instructed to inspect changes after verification results, accumulated findings or new strategic guidance, then propagate statuses and re-score affected nodes. A node is mechanically marked stale when an incident edge has a timestamp later than its current annotation, making newly connected evidence visible as a reason to reconsider the score. The frontier includes open or active plans, directions and annotated symbolic goals. Linked nodes rank ahead of unlinked nodes; within each group, scored nodes precede unscored nodes, higher scores rank first and remaining ties favour older records. Unlinked proposals are left unscored under the curator contract until workers connect them to the graph. Workers consult this frontier when choosing the next step within their assignments.

The curator's status rules follow the structure of the proposed argument. An AND route is complete only when all its obligations, including any reassembly step, are complete; a failed obligation invalidates that route. Alternative routes remain independent. A reduction from $P$ to $Q$ supports completion of $P$ only when $Q$ is complete and an accepted bridge establishes $Q\Rightarrow P$. Equivalence requires supporting facts in both directions, whereas a proved generalization supports its special case through a specialization fact. Refutation of the special case also refutes the generalization. Obstacles and counterexamples justify terminal or blocked statuses only when verified or otherwise decisive. These are instructions applied by the curator to the recorded evidence; the storage layer validates relation types, statuses and score ranges without performing mathematical inference.

Scoring combines an assessment of potential progress with a heuristic exploration bonus for less-explored directions. The curator also considers how many accepted facts directly depend on a prerequisite, so that a result supporting several ongoing arguments can receive greater attention. Strategic guidance takes precedence over the ranking, and disagreements are recorded in comments. No fixed numerical exploration-bonus formula or trained scoring model is imposed. Scores prioritize further work, while a \code{closed} annotation records an exploration judgment; neither operation admits a mathematical fact.

\subsection{Graph-based retrieval}
Graph-based retrieval combines lexical search with the proof dependencies and exploratory relationships recorded in project memory (Fig.~\ref{fig:cgm}d). Agents first locate relevant findings or facts using BM25. Search ranks complete shared-finding records and stored fact documents. Finding results contain metadata and claim previews, after which agents request selected records in full to inspect their evidence, verification feedback and current status. Fact search returns identifiers and statements that provide entry points for neighbourhood retrieval.

For a selected fact, neighbourhood retrieval follows two complementary structures. The fact graph supplies its declared predecessors, exposing the premises on which the argument depends. The exploration graph supplies additional accepted facts connected directly to the selected fact or through an intermediate non-fact node within two steps. For example, two facts linked to the same symbolic goal can be retrieved together even when their statements share no query terms. Traversal considers either edge direction, while returned paths preserve the relation labels and directions so that the agent can interpret each connection. Repeated facts reached through multiple exploration paths are deduplicated, and facts already returned as predecessors are excluded from the additional results.

Neighbourhood retrieval prioritizes declared predecessors before returning additional facts connected through the exploration graph. Immediate dependents can also be requested, and their count is always reported to indicate how many accepted arguments directly rely on the selected fact. Results provide statements and connecting paths; agents read the full proofs of relevant facts separately. Search and neighbourhood views are reconstructed from the current stores, making newly admitted facts and newly recorded relations available to subsequent queries. This mechanism assembles relevant research context from accumulated exploration trajectories, while keeping exploratory connections distinct from declared proof dependencies.

\subsection{Cross-project experience and continual memory}
Cross-project experience makes selected knowledge from earlier exploration trajectories available to a new investigation (Fig.~\ref{fig:cgm}e). A deployment setting disables this access, restricts it to projects with the same batch label or allows access across projects. The retrieval filter enforces project membership; the run protocol determines which completed projects are available in a sequential study. Workers and the main agent can query three channels: accepted fact statements, negative findings comprising dead ends, obstacles and counterexamples, and retrospective curator lessons. Strategic guidance and progress summaries are excluded from this recall interface.

Each channel is ranked independently with BM25, and the requested result budget is divided across channels. Results retain source-project provenance, with record types, timestamps or fact identifiers as applicable. Fact recall returns statements; negative-finding recall returns claims and short evidence excerpts; lesson recall returns a scoring pattern and its proposed adjustment. These records can suggest relevant results, identify previously encountered obstacles or guide the choice of an approach. A retrieved mathematical claim must be proved and admitted within the receiving project before it can serve as a local premise.

Lessons provide a second route for transferring experience. At project close, the curator can review its annotation history against recorded route outcomes and identify recurring scoring errors. Each lesson stores a pattern, supporting evidence and a concrete adjustment, together with its source project and author, in a deployment-level log. The contract restricts lessons to methodological guidance, such as requiring a separately established reassembly step before assigning a high score to a localization plan. The launcher adds the latest 20 eligible lessons to the next curator's initial context under the same visibility setting. Experience therefore changes the information available for subsequent search and curation through external memory, while model parameters and the curator's standing contract remain fixed.

The formal memory specification and execution pseudocode are provided in Supplementary Note~3. Benchmark configurations, grading procedures, re-review selection and resource-accounting conventions are detailed in Supplementary Note~7.

\begin{figure}[t]
\centering
\includegraphics[width=\linewidth,height=0.67\textheight,keepaspectratio]{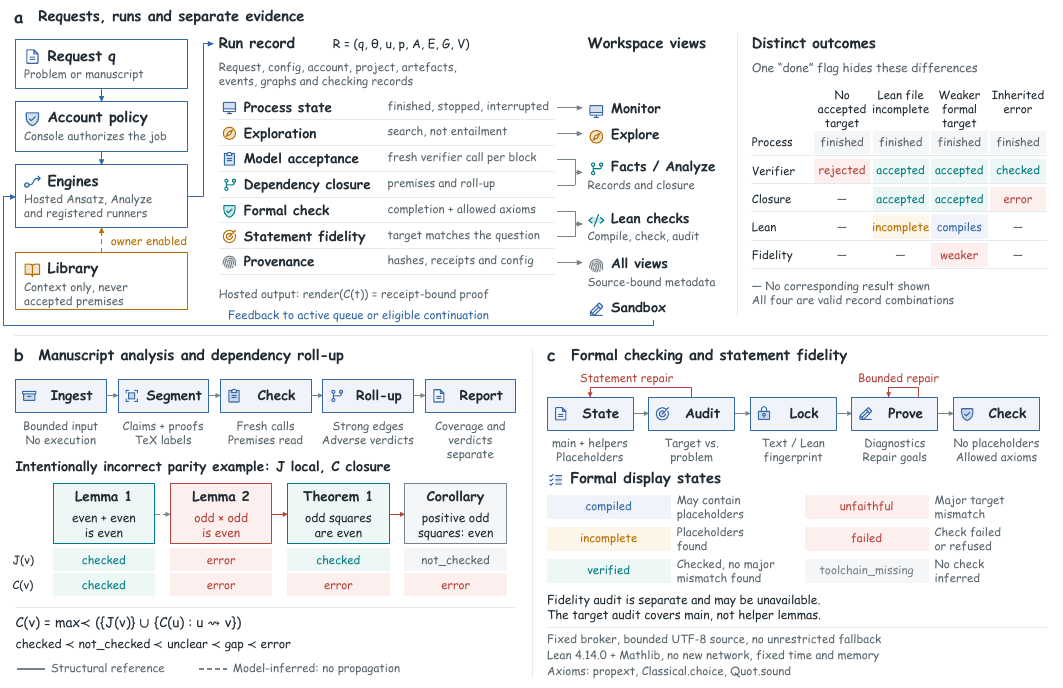}
\caption{\textbf{Platform evidence model and checking workflows.} \textbf{a},~Requests pass through account policy and engines into run records whose execution, exploration, model acceptance, dependency closure, formal checks, statement fidelity and provenance remain separately inspectable. Library content supplies context only; hosted accepted output is matched to the receipt-bound rendering of the accepted closure. The four examples show combinations obscured by a single success flag. \textbf{b},~Manuscript analysis reviews results individually and propagates adverse verdicts along structurally grounded dependencies. In the intentionally incorrect parity example, Theorem~1 is locally checked conditional on its premises but inherits Lemma~2's error; the unchecked corollary inherits the same error. Dashed, model-inferred links do not support this propagation. \textbf{c},~Formalization audits and locks the target before bounded proof construction and source checking. Compilation, placeholder-free checking and statement fidelity have different scopes. The displayed \code{verified} state reports successful completion and soundness checks with no major mismatch detected by available fidelity checks; an unavailable audit does not imply a positive fidelity judgement. The audit concerns \code{main}, not every helper lemma. Schematics illustrate implementation contracts rather than model accuracy.}
\label{fig:platform_evidence}
\end{figure}

\subsection{Community workspaces and hosted execution}
\Ansatze\ exposes research projects as shared workspaces linking requests, configurations, artefacts, graph states and checking records~\cite{platformrepo}. Users can inspect arguments, edit proofs, provide feedback and continue an investigation from its retained state. Concurrent edits are checked before saving to avoid overwriting intervening changes. Catalogue entries preserve source attribution, while atlas links record correspondences between informal and formal material. Public research-memory admission requires an eligible completed run, owner authorization, an atlas association and model provenance, with additional checking and fidelity requirements for formal records. Retrieved library material supplies context and is not automatically admitted as a premise; forum discussion is excluded from harness context selection. Execution progress is displayed separately from evidential status, so a completed run can still contain unresolved claims.

The hosted controller provides a bounded execution path through scaffolding, exploration, verification, curation and synthesis. Model-backed stages use response-only calls. Application checks establish required fields, valid predecessors and an acyclic dependency structure. Verification then follows the target's predecessor closure, reviewing each argument with its accepted premise statements. Critical errors or substantive gaps prevent acceptance; oversized inputs are refused, and repair proceeds within reserved budgets. Publication renders the accepted target and its supporting closure. A controller receipt binds this rendering to the exploration records, calls and configuration, providing provenance within the trusted host boundary~\cite{provdm}.

\subsection{Manuscript analysis and formal checking}
Manuscript analysis examines a supplied argument as written. Ingestion bounds archive expansion, checks entry paths and network destinations, and skips computational material without executing it. Segmentation identifies statements, proofs, source locations and dependencies, distinguishing structural references from model-inferred associations. Fresh reviews record successful checks, gaps, errors, uncertainty or absence of review. Uninterpretable responses are excluded from reviewed coverage. Adverse judgments propagate through structurally grounded dependencies, allowing a locally checked argument to inherit an upstream error. Model-inferred links support inspection without receiving the same dependency interpretation. Optional Lean output supplies statement skeletons.

Formalization separates translation of the intended problem from proof construction and checking~\cite{lean4,autoformalization}. A proposed target is elaborated and audited for changed domains, added hypotheses, omitted cases or weakened conclusions. Textual and elaborated target locks constrain later revisions, while diagnostic feedback guides bounded repair. Checking additionally detects placeholders and inspects declaration axioms under the permitted policy~\cite{leanvalidation}. The checking service binds evidence to the source and toolchain. Statement-fidelity auditing may remain unavailable even after a successful formal check. A successful check does not establish that the initial translation was faithful, and a main-target audit does not cover every helper; source changes can invalidate earlier evidence.

Platform implementation checks use isolated state and synthetic dependencies without paid model calls. Historical checking replays and scripted interface fixtures document software behaviour rather than community outcomes. The published source of the platform's formal example was matched to the reported checksum, but its historical prover execution was not repeated during synthesis, and the separate raw validation bundle was absent from the supplied archive. Detailed configurations, test records and evidence limitations are provided in the Supplementary Information.

\begin{figure}[t]
\centering
\includegraphics[width=\linewidth,height=0.67\textheight,keepaspectratio]{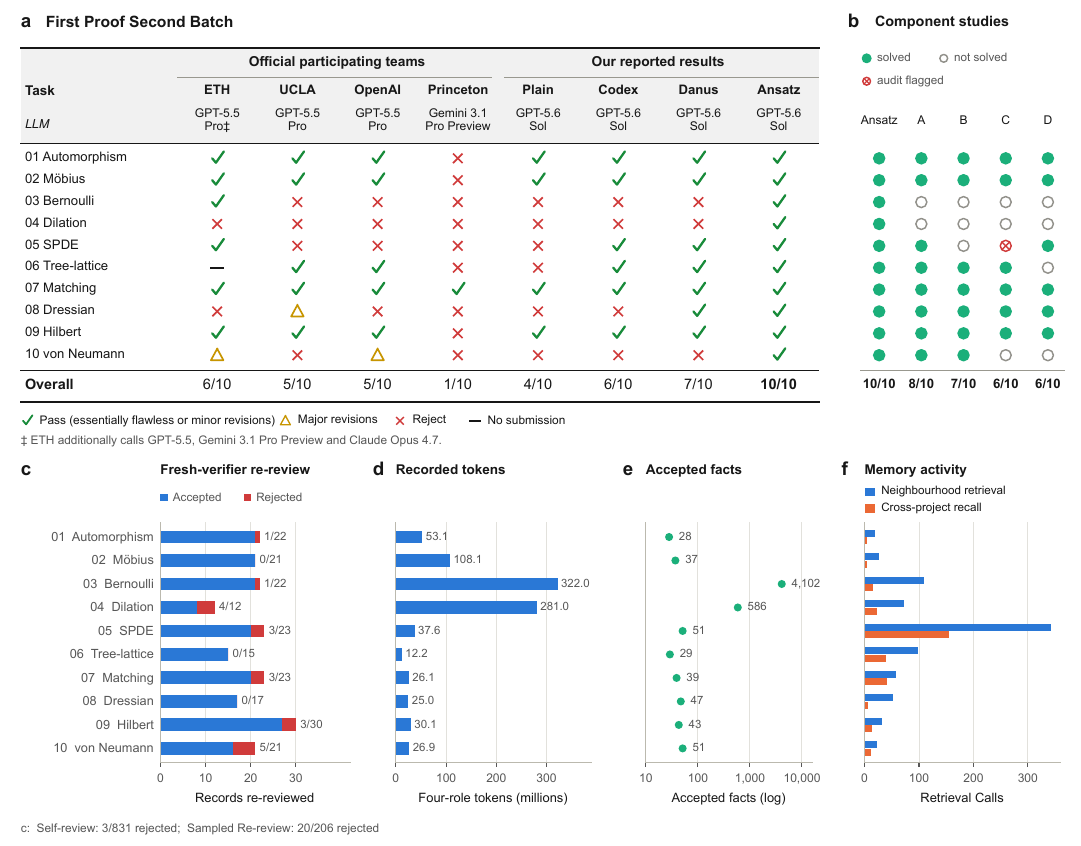}
\caption{\textbf{First Proof Second Batch results, ablation studies and research activity.} \textbf{a},~Problem-level outcomes for four official participating teams and four reported systems. Check marks indicate passes under the respective criteria; triangles, major revisions; crosses, rejection; dashes, no submission. 
Ansatz recorded ten passes under the study's grading criterion: nine full-target model-verifier closures and one partial Bernoulli result.
\textbf{b},~Ablation studies: A, flat fact notebook without fact graph; B, no exploration graph or curator; C, worker self-review without a separate verifier; D, no strategy consultation. Green and open circles denote solved and not solved under the study's criterion; crossed circles flag cases initially marked solved but rejected by subsequent target or dependency review and excluded from the totals. 
\textbf{c},~Fresh-verifier re-review of self-accepted records for ablation study C, with blue and red bars indicating acceptance and rejection. The audit covers all available records for tasks 01, 02, 06 and 07, and named targets plus 12 records selected by a fixed stride for each remaining task. Bar-end labels report rejected/re-reviewed counts.
The footer gives the self-review and sampled re-review totals. \textbf{d},~Recorded tokens across workers, verifier, main agent and curator.
\textbf{e},~Accepted facts in each project's latest snapshot, on a logarithmic scale. 
\textbf{f},~Neighbourhood-retrieval and cross-project-recall calls.}
\label{fig:audit}
\end{figure}

\section{Results}

\subsection{First Proof Second Batch}
First Proof Second Batch comprises ten research-level mathematics problems contributed by mathematicians across fields~\cite{firstproof2}. Figure~\ref{fig:audit}a compares published results for four official participating teams with the reported results for Plain, Codex, Danus~\cite{danus} and Ansatz. Ansatz recorded a pass on all ten problems under the study's grading criterion, comprising nine full-target model-verifier closures and a partial Bernoulli characterization. The corresponding reported totals for Plain, Codex and Danus were four, six and seven passes.

The problem-level results show where this additional coverage arose. All four reported systems passed automorphism, M\"obius, matching and Hilbert. Codex and Danus additionally passed SPDE and tree lattice, and Danus also passed Dressian. Ansatz retained these seven passes and added Dilation, von Neumann and the Bernoulli partial result. Dilation and von Neumann received no passes from any of the other systems shown. Bernoulli was passed by ETH among the official teams, and Ansatz's partial result also counted as a pass under the study's criterion.

These results highlight the breadth of Ansatz's reported coverage, including progress on problems that remained unresolved by the other systems shown. Its persistent workspace allows investigations to continue from accumulated findings, supporting sustained work on difficult targets.

\subsection{Ablation Studies and Additional Analysis}
We examined four ablations of a common five-worker configuration (Fig.~\ref{fig:audit}b): replacing structured facts with a flat notebook (A), removing the exploration graph and curator (B), substituting worker self-review for the separate verifier (C), and disabling strategy consultation (D). The retained outcomes were eight, seven, six and six passes, respectively, compared with ten for the displayed Ansatz reference. All four arms retained automorphism, M\"obius, matching, Dressian and Hilbert, while none solved Bernoulli or Dilation. Their differences were concentrated in three tasks: SPDE was retained by A and D, tree lattice by A--C, and von Neumann by A and B. These patterns identify task-specific sensitivities to the ablated configurations. Ansatz brings these capabilities together in a shared workspace, allowing search, review and strategic guidance to operate on accumulated research state.

Re-review exposed a further distinction between reaching a target and retaining it after another assessment (Fig.~\ref{fig:audit}c). Worker self-review rejected three of 831 submissions, whereas a fresh verifier rejected 20 of 206 selected self-accepted records. The audit flagged the SPDE result, reducing C from seven problems initially marked solved to six after re-review. Selection combined complete review of four smaller graphs with targeted and fixed-stride sampling of larger graphs. This audit illustrates how a separate reviewer can challenge arguments accepted by proposing workers. Recording those judgements alongside the corresponding claims makes correction and subsequent scrutiny part of the research workflow.

Resource use was concentrated in the most persistent investigations (Fig.~\ref{fig:audit}d). Bernoulli and Dilation used 322.0 and 281.0 million recorded tokens, together accounting for 65.4\% of the 922.0-million total. These records illustrate Ansatz's capacity to sustain extended investigations and to expose their computational demands across agent roles.

Bernoulli also contributed 4,102 of the 5,013 accepted facts, yet remained partial; Dilation was solved with 586 facts (Fig.~\ref{fig:audit}e). Thus, the amount of accumulated research state did not determine whether the full target was solved. By retaining intermediate statements and their dependencies, Ansatz preserves a basis for continuing or redirecting investigations whose targets remain unresolved.

For retrieval, 834 neighbourhood-retrieval calls and 315 cross-project-recall calls were recorded (Fig.~\ref{fig:audit}f). SPDE accounted for the largest counts in both channels, with 343 and 155 calls, respectively. These records show where memory was consulted, while establishing successful mathematical reuse requires tracing retrieved material into subsequent arguments. The two retrieval channels give agents access both to nearby project evidence and to experience from earlier investigations, making accumulated research accessible during ongoing search.

\begin{figure}[t]
\centering
\includegraphics[width=\linewidth,height=0.67\textheight,keepaspectratio]{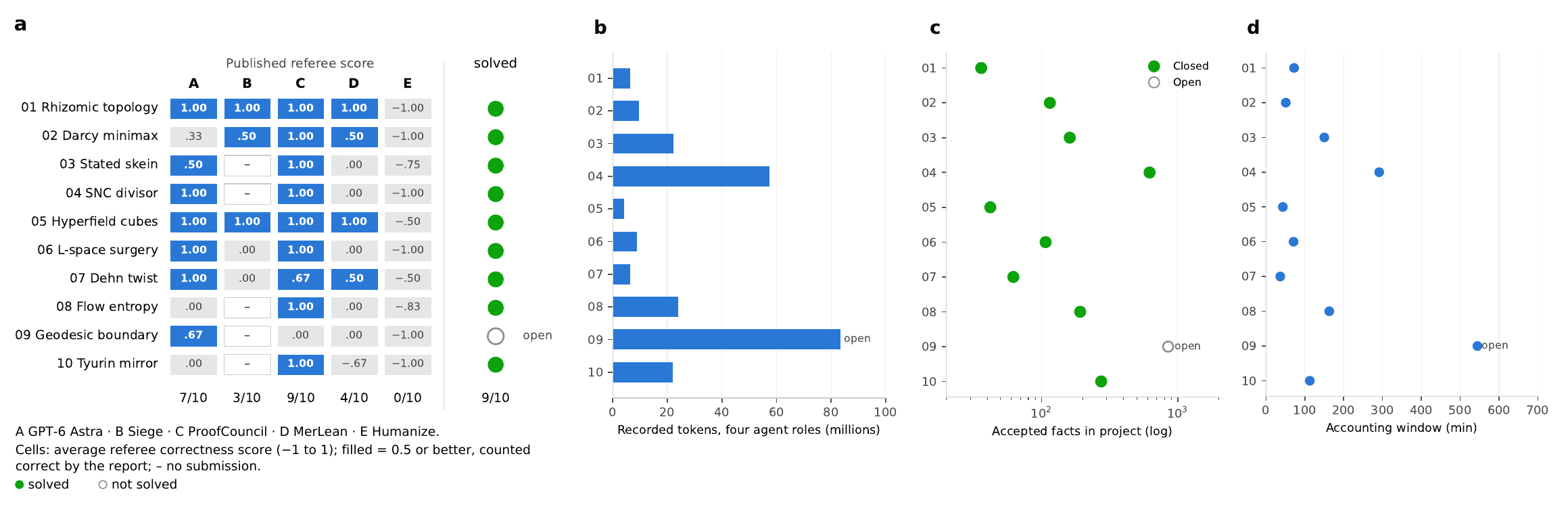}

\caption{\textbf{Evaluation of the harness on First Proof Third Batch.}
\textbf{a},~Published referee correctness scores for five external systems
(A, GPT-6 Astra; B, Siege, a Claude Opus 5.5 workflow; C, ProofCouncil;
D, MerLean; E, Humanize, open-weight). Each cell is the average referee
score on the report's scale from $-1$ to $1$; filled squares score $0.5$
or better, the report's threshold for a correct submission; --, no
submission. Beside them, local closures of the harness (filled circles:
the verbatim target became a fact accepted by the deployment's own
language-model verifier; closures resting on cited published theorems or
declared conventions are counted as closed). The two endpoints differ and
are not comparable. \textbf{b},~Recorded tokens of the four agent roles
per problem. \textbf{c},~Accepted facts per project. \textbf{d},~Accounting
windows. For problems 4 and 9 the run comprises the capped search and its
continuation with the time cap lifted. All values are read from the
archived records; no expert has graded any local closure.}

\label{fig:fpb3}
\end{figure}

\subsection{First Proof Third Batch}

Ansatz recorded local verifier closures on nine of the ten First Proof Third Batch problems (Fig.~\ref{fig:fpb3}a). Among the external systems, ProofCouncil reached the published referee-score threshold of 0.5 on nine problems, followed by GPT-6 Astra on seven, MerLean on four, Siege on three and Humanize on none. The local closures occurred on the same nine problems on which ProofCouncil reached this threshold. These counts reflect different assessment procedures and do not establish a ranking under a common evaluation protocol.

The problem-level comparison reveals variation obscured by the aggregate counts. Rhizomic topology and hyperfield cubes received full referee scores from four external systems and also reached local verifier closure in Ansatz. Flow entropy and Tyurin mirrors were more discriminating: ProofCouncil was the only external system to reach the scoring threshold on either problem, and Ansatz recorded local closures on both. Ansatz also recorded a local closure on Darcy minimax, on which GPT-6 Astra scored below the threshold. Conversely, geodesic boundaries remained unclosed by Ansatz and fell below the threshold for ProofCouncil, although GPT-6 Astra received a score of 0.67. Thus, the local-closure and referee-score records identify a shared remaining gap on a problem where another system met the benchmark's scoring criterion.

Recorded computation varied substantially across problems (Fig.~\ref{fig:fpb3}b). Geodesic boundaries consumed approximately 83 million tokens across the four agent roles, the largest recorded total, followed by SNC divisors at approximately 57 million. Each remaining problem used fewer than 25 million tokens. The contrast between the two largest runs is informative: Ansatz recorded a local verifier closure on SNC divisors, whereas geodesic boundaries remained unclosed despite greater recorded token use.

The accepted-fact counts and accounting windows show a corresponding concentration of research activity (Fig.~\ref{fig:fpb3}c,d). Geodesic boundaries accumulated the largest fact graph and the longest accounting window, approximately 550 minutes; SNC divisors had the next-largest graph and a window of approximately 290 minutes. The other local closures were reached with substantially smaller graphs and shorter windows. These observations distinguish the accumulation of accepted intermediate statements from solving the final target: extensive recorded activity and a large research graph did not, by themselves, ensure a solution.

\begin{table}[t]
\centering\small
\caption{\textbf{Ansatz on open problems.} Full denotes a result addressing the open target; Partial denotes a strict improvement that leaves the target unresolved. Status is recorded separately: Lean denotes an author-reported formalization in Lean of the reported result; AI-proposed denotes a result accepted by the harness's model-based review that has not been formalized in Lean. Numbers refer to the Erd\H{o}s problems catalogue~\cite{bloomerdos}; statements and results are given in \hyperref[sec:si-open-problems]{Supplementary Note~9}.}
\label{tab:open}
\renewcommand{\arraystretch}{1.15}
\setlength{\tabcolsep}{4pt}
\begin{tabularx}{\linewidth}{@{}L{2.9cm}YYL{2.05cm}lL{1.65cm}@{}}
\toprule
Problem & Previously known & Ansatz result & Open target & Claim & Status\\
\midrule
Jamison caterpillar conjecture~\cite{Jamison83,Jamison84} & Verified for $n\le24$ (ref.~\citenum{MolOellermann19}); open in 2026 (ref.~\citenum{Wang26}) & Every maximizer of the mean subtree order is a caterpillar, for all $n$ & All $n$ & Full & Lean\\
561, size Ramsey numbers of star forests~\cite{BEFRS78} & Equal sizes; one star; two equal stars; all sizes odd~\cite{DJKR25} & The conjectured formula for every pair of star forests & All pairs & Full & AI-proposed\\
289, reciprocal sums over separated intervals~\cite{ErGr80} & Open; unrestricted variant settled~\cite{Kovac289} & A threshold $k_0$ exists (not effective) & All large $k$ & Full & Lean\\
488, density of multiples & Open & Counterexample with $d_A(m)/d_A(n)>256/67$ & Decide the factor-2 bound & Full & Lean\\
348, deletions from complete sequences~\cite{ErGr80} & $m=0,1$ possible & Exactly $m\in\{0,1\}$ (eventual completeness) & Classify all $m<n$ & Full & Lean\\
Conway's thrackle conjecture~\cite{LPS97} & $|E|\le1.393(|V|-1)$ (ref.~\citenum{Xu21}); $1.385(|V|-1)$ for $|V|\ge4$ (preprint~\cite{Scoppetto26}) & $|E|\le29|V|/22$ for every thrackle & $|E|\le|V|$ & Partial & AI-proposed\\
Graceful tree conjecture~\cite{Ringel64,Rosa67} & At most four leaves~\cite{HKR82}; spiders with at most four legs of length $\ge2$ (ref.~\citenum{PaPo16}); diameter $\le5$ (ref.~\citenum{HrHa01}) & Every tree with at most seven leaves; every spider with at most eight legs of length $\ge2$ & All trees & Partial & AI-proposed\\
811, rainbow subgraphs of balanced colourings & Most cliques; $K_4$ (refs~\citenum{AxCl24,ClWa23}) & No clique $K_q$ with $q\ge4$ is forced & All graphs & Partial & AI-proposed\\
644, transversals of $(7,2)$ families & $\lceil7k/8\rceil$ (ref.~\citenum{FDFKW00}) & $\lceil(6k+3)/7\rceil+1$ for $k\ge150$ & Order $3k/4$ & Partial & AI-proposed\\
993, unimodality of independence sequences & Increasing to about $n/8$ for trees~\cite{BaGa21} & Strictly increasing to $131n/520$ for forests & Unimodality & Partial & AI-proposed\\
1063, binomial divisibility & Subexponential upper bound claimed~\cite{EPAD1063} & Super-polynomial lower bound; refined upper bound & Sharp growth of $n_k$ & Partial & Lean\\
302, reciprocal-sum-free sets~\cite{ErGr80} & Coefficients above $5/8$ and below $0.860853$ (preprint~\cite{Khanukov26}) & $f(N)/N\to c$; explicit gain over $5/8$ & Determine $c$ & Partial & AI-proposed\\
\bottomrule
\end{tabularx}

\end{table}

\subsection{Discoveries on longstanding open mathematical problems}

We applied Ansatz to twelve longstanding open problems in combinatorics and number theory, drawn from the Erd\H{o}s problems catalogue and the mathematical literature~\cite{bloomerdos,Jamison83,LPS97}. The investigations produced five manuscripts claiming complete resolutions of their respective open targets, alongside seven projects reporting partial advances (Table~\ref{tab:open}). These results span extremal graph theory, Ramsey theory, combinatorial number theory and related areas, illustrating the potential of persistent, multi-agent mathematical exploration to pursue research questions beyond established benchmarks.

Among the full-target claims, two address longstanding conjectures in graph theory. The Jamison caterpillar conjecture, introduced in the 1980s, asserts that every tree maximizing the mean subtree order among trees with $n$ vertices is a caterpillar~\cite{Jamison83,Jamison84}. Previously verified for $n\le24$ (ref.~\citenum{MolOellermann19}) and still described as open in 2026 (ref.~\citenum{Wang26}), the conjecture is addressed by a manuscript proposing a proof for all $n$. The argument reduces a shortest terminal arm of a maximizing tree to a broom configuration and uses exact polynomial certificates to establish the required transformation. For Erd\H{o}s problem 561, Ansatz generated a manuscript claiming the conjectured size Ramsey formula for every pair of star forests, extending beyond previously established special cases~\cite{BEFRS78}. Both investigations exemplify how extended proof search can develop arguments toward problems that have remained unresolved for decades.

Three further manuscripts address open questions in combinatorial number theory. For Erd\H{o}s problem 289, Ansatz constructs pairwise disjoint, non-adjacent intervals whose reciprocal sums equal one for every sufficiently large number of intervals, using a Pell-type construction, equal-mass packets and a summable reserve. The resulting existence threshold is ineffective, and a related public manuscript establishes a stronger bounded-length formulation~\cite{Tang289}. For problem 488, the proposed construction disproves the conjectured factor-two bound on densities of multiples by exhibiting a finite set $A$ and admissible integers $m>n\ge\max A$ satisfying $d_A(m)/d_A(n)>256/67$. For problem 348, the manuscript characterizes all deletion pairs preserving eventual completeness, identifying precisely $(0,n)$ for $n\ge1$ and $(1,n)$ for $n\ge2$.

Beyond these five full-target claims, seven additional investigations produced candidate advances without resolving their complete targets. These include an improved transversal bound for Erd\H{o}s problem 644, from $\lceil7k/8\rceil$ to $\lceil(6k+3)/7\rceil+1$ for $k\ge150$~\cite{FDFKW00}, and a result covering all clique cases of the rainbow-subgraph question in problem 811~\cite{AxCl24,ClWa23}. Further projects investigate independence sequences, reciprocal-sum-free sets, binomial divisibility, Conway's thrackle conjecture and graceful tree labellings. These partial results are important to the research framework because useful mathematical progress need not take the form of a complete proof, such as improved bounds, where new constructions and intermediate reductions can provide reusable starting points for subsequent investigations. To measure the correctness, verification was pursued at multiple levels. The source projects report Lean formalizations of the results for the Jamison caterpillar conjecture and Erd\H{o}s problems 289, 348, 488 and 1063. The remaining seven projects were assessed through the harness's model-based review, with preliminary feedback from mathematicians reported for some results. 

Taken together, these investigations illustrate the broader research potential of Ansatz: rather than terminating at an unsuccessful proof attempt, the system can preserve mathematical constructions, partial arguments and intermediate advances for continued exploration and scrutiny. The available manuscripts document the proposed results, although the corresponding open-problem execution and retrieval records were not included in the benchmark archive. We therefore present these findings as candidate mathematical contributions, with available manuscripts released for community examination~\cite{cgmrepo}. Their broader significance lies both in the mathematical questions they address and in demonstrating a research workflow designed to sustain, preserve and expose progress on problems that may require extended investigation.

\begin{figure}[t]
\centering
\includegraphics[width=\linewidth,height=0.67\textheight,keepaspectratio]{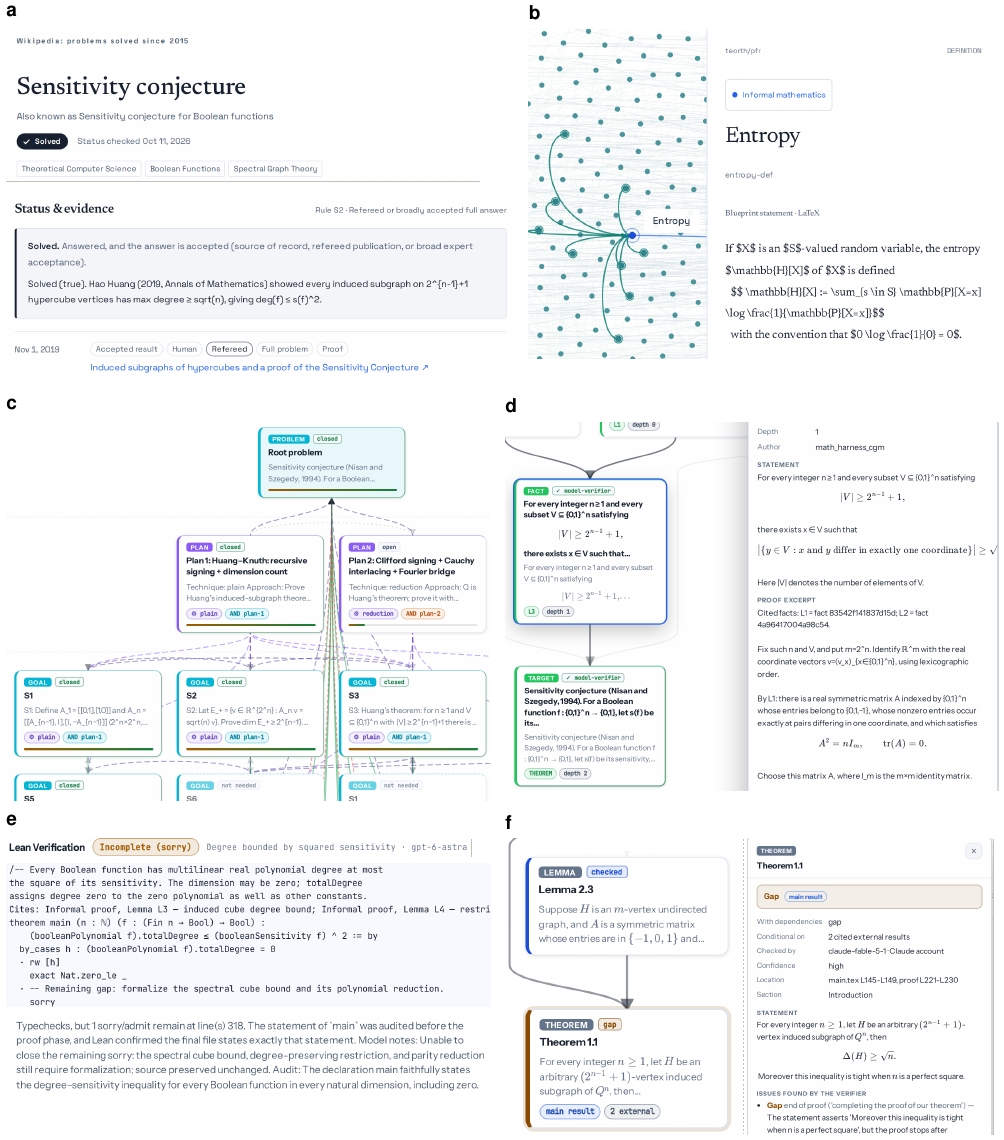}
\caption{\textbf{Platform views expose research state and checking outcomes.} \textbf{a},~Catalogue entry for the sensitivity conjecture, showing its solved status and supporting source. \textbf{b},~Atlas view centred on entropy, linking a graph entry to its definition. \textbf{c},~Explore graph with plans, goals and their recorded states for a sensitivity-conjecture investigation. \textbf{d},~Facts graph and a selected record, with its statement, identifier and proof. \textbf{e},~Lean view for a bound on degree by squared sensitivity, displaying \code{Incomplete (sorry)} alongside checking and fidelity notes. \textbf{f},~Analyze view linking Lemma~2.3, labelled \code{checked}, to Theorem~1.1, labelled \code{gap}, with the theorem's review details. These views distinguish search progress, model review and formal-check outcomes; they do not by themselves establish correctness of a complete result.}
\label{fig:ui}
\end{figure}

\subsection{A community workspace for inspecting and extending research}

Beyond Ansatz, \Ansatze\ extends persistent mathematical investigation into a collaborative research environment where researchers can inspect, evaluate and build upon AI-generated work. Unlike conventional AI chat interfaces, which primarily present mathematical reasoning as sequential conversations, \Ansatze\ provides a structured, interpretable and accessible representation of the research process. By organizing complex arguments, intermediate findings and their dependencies into interconnected research graphs, the platform helps mathematicians navigate lengthy investigations, understand the reasoning behind proposed results and critically assess their validity. Each investigation preserves its original request, execution configuration, project context, artifacts, event history, research graphs, and checking records (Fig.~\ref{fig:platform_evidence}), allowing researchers to trace the development of mathematical ideas, revisit alternative approaches, and extend previous work without losing its underlying context.

To make this research process transparent and accessible, the workspace provides complementary views of ongoing investigations (Fig.~\ref{fig:ui}). The Explore view visualizes research plans, intermediate findings and obstacles, while the Facts view presents admitted mathematical claims and their declared dependencies. Execution monitoring tracks agent activity, providing insight into how investigations progress and how results are obtained. Hosted results are linked to receipt-bound representations of their accepted dependency closures, preserving provenance without implying independent mathematical correctness. Moreover, eligible completed investigations can contribute to shared research memory, enabling researchers and agents to retrieve and build upon previous findings while requiring mathematical claims to be independently established before their reuse as verified premises.

Beyond research exploration, \Ansatze\ facilitates deeper mathematical understanding and collaborative scrutiny through manuscript analysis, research communication and formal verification. Manuscript analysis evaluates arguments individually and traces how gaps or errors in supporting premises affect downstream conclusions, helping researchers identify precisely which parts of a proof require further examination. The platform also supports transforming AI-generated mathematical investigations into coherent, human-readable manuscripts, organizing complex reasoning, and transferring intermediate discoveries into structured mathematical narratives. This capability helps bridge the gap between autonomous AI reasoning and human mathematical understanding, enabling mathematicians to interpret, evaluate and extract insights from AI-generated results. Finally, \Ansatze\ integrates Lean-based formalization of proposed proofs, providing additional checks on proof completeness, permitted axioms and fidelity to the intended mathematical statement~\cite{lean4,leanvalidation,autoformalization}. Together, these capabilities connect autonomous mathematical exploration with human interpretation, critical evaluation and formal checking, supporting a more transparent and collaborative approach to AI-assisted mathematical discovery.

\FloatBarrier
\section{Discussion}
Mathematical discovery requires a research process that can preserve useful work while its direction changes. Ansatz addresses this requirement by representing both mathematical results and the evolving exploration that produces them. Its exploration graph records plans, findings, obstacles, counterexamples and other intermediate results of exploration alongside a fact graph of accepted statements and their dependencies. Continual memory then makes selected knowledge from these trajectories available to subsequent exploration. The benchmark and open-problem investigations illustrate how this architecture supports extended work across sessions and retains progress even when a target remains unresolved. Together with the \Ansatze\ platform, it provides a persistent workspace through which agents and researchers can inspect, resume and extend mathematical investigations.

The exploration graph is particularly important for sustaining long-horizon mathematical exploration. A difficult investigation can involve several competing reductions, repeated changes of strategy and intermediate results whose relevance becomes apparent only later. By preserving the relationships among these activities, the graph gives workers access to the reasoning context surrounding a finding: which approach produced it, what obstacle it addresses and which directions remain open. Curator annotations help organize this evolving research frontier, while neighbourhood retrieval brings related facts and their exploratory connections into the current context. Persisted project state also allows new worker sessions to recover relevant progress and continue an unfinished line of inquiry. The differing ablation outcomes are consistent with a role for these functions in supporting search, while the continued runs directly demonstrate persistence across sessions. We therefore view explicit exploration state as a practical mechanism for maintaining continuity through extended mathematical investigations.

Building on fact-graph research agents~\cite{danus}, Ansatz expands the reusable research record to include the trajectories through which candidate results were developed. Continual memory connects earlier and current exploration by retrieving mathematical statements, negative findings and methodological lessons. An earlier counterexample can suggest a missing hypothesis, an unsuccessful reduction can help identify an obstacle, and a useful proof strategy can motivate a new approach. This gives prior exploration an ongoing role in the research process, including work that did not produce a complete solution. Such adaptation occurs through external memory, with model parameters fixed within a deployment. Retrieved mathematical statements require local proof and admission, while methodological lessons can guide the next investigation. The retrieval records establish that these channels are used during search; their intended benefit is to help agents make better use of accumulated experience. 

Beyond supporting autonomous mathematical exploration, \Ansatze\ provides a bridge between AI-generated research and human mathematical understanding. Unlike conventional chat-based interfaces, which often present lengthy investigations as sequential interactions, the platform organizes mathematical claims, intermediate findings, proof dependencies and exploration histories into structured, navigable representations. This enables researchers to examine how conclusions were reached, identify gaps in reasoning and build upon previous investigations without reconstructing the entire research process. By integrating manuscript analysis, human-readable research communication and Lean-based formalization, \Ansatze\ also supports the transition from autonomous proof search to mathematical results that researchers can interpret, scrutinize and refine. More broadly, the platform offers a foundation for collaborative human--AI mathematical discovery, in which AI agents contribute sustained exploration while mathematicians provide critical judgement, conceptual insight and independent validation. For future work, we will hope to focus on continuously improving the platform's accessibility, interpretability and collaborative capabilities, with the goal of making AI-assisted mathematical research more efficient, reliable and accessible to the broader mathematical community.

As we look beyond solving individual mathematical problems, an important challenge is to make sustained AI-assisted research more efficient, accessible and useful to mathematicians. Reducing computational costs and improving the interpretability of AI-generated reasoning will be essential to enabling broader participation in mathematical discovery. Ultimately, we envision AI evolving from a tool for solving isolated problems into a collaborative research partner that supports mathematicians in developing new ideas, exploring deeper connections and advancing mathematical knowledge.

\section{Conclusion}
We have presented Ansatz and \Ansatze\ as a connected framework for cumulative mathematical research. Ansatz organizes accepted claims and their dependencies alongside an evolving record of plans, obstacles and partial results, allowing parallel agents to preserve and revise useful work across extended investigations. On the First Proof benchmark, the harness recorded nine local verifier closures in each batch, with the partial Bernoulli result also counting as a Second Batch pass and partial progress on the remaining Third Batch problem. Applied to open problems, it produced results addressing five full targets and advancing seven others; five Lean formalizations were reported, while the remainder are AI-proposed. The component studies show why persistent memory needs explicit admission and revocation rules: errors accepted during self-review can expose later dependent arguments. The \Ansatze\ platform makes these research records available for inspection and feedback, presenting model review, dependency closure, formal checks and statement fidelity separately. Independent expert review of the benchmark and AI-proposed results, controlled repeated evaluations and complete provenance records remain necessary to establish the reliability and generality of this approach~\cite{provdm,stodden2016,fair}.

\nolinenumbers

\begingroup
\footnotesize
\setlength{\bibsep}{1pt}

\endgroup

\clearpage
\section{Extended Data}
\begingroup
\setcounter{table}{0}
\renewcommand{\theHtable}{ED\arabic{table}}
\captionsetup[table]{name=Extended Data Table}
\begin{table}[ht]
\centering\small
\caption{\textbf{Evidence types recorded by the platform and harness.} Each type answers a different question; none implies another.}
\label{tab:evidence}
\renewcommand{\arraystretch}{1.18}
\begin{tabularx}{\linewidth}{@{}L{2.55cm}L{3.0cm}YY@{}}
\toprule
Evidence type & Produced by & Establishes & Does not establish\\
\midrule
Process state & Job manager or worker loop & Which stages ran and whether they finished, stopped or were interrupted & That the mathematical goal was resolved\\
Model-verifier acceptance & Fresh verifier call or session per block or fact & The submitted argument passed a model review given the cited premise statements & Correctness, statistically independent error or expert endorsement\\
Dependency closure & Parser, fact store, Analyze roll-up & Declared premises, the admitted closure and inherited adverse verdicts & That every premise used was declared, or global consistency\\
Exploration record & Workers and curator & Plans, findings, obstacles, counterexamples and other intermediate results of exploration with typed relations and curator status & Entailment, or the admission of any fact\\
Formal check & Fixed-interface Lean checker & The exact source elaborates without placeholders and uses only allowed axioms & That the formal target expresses the intended question\\
Statement fidelity & Separate model audit, statement locks & No weakened conclusion, omitted case or extra hypothesis was detected; target locked across edits & Faithfulness of the original translation, or of helper lemmas\\
Provenance & Hashes, receipts, configuration and accounting records & Correspondence between a displayed artefact and the records and interventions that produced it & Reproducibility of stochastic model output, or integrity after host compromise\\
External review & Experts, outside the system & Community judgement of correctness, novelty and significance & Not produced by the harness\\
\bottomrule
\end{tabularx}
\end{table}
\endgroup

\fi
\ifsupplementpart
\clearpage
\nolinenumbers
\newgeometry{margin=2.2cm}
\setcounter{page}{1}\renewcommand{\thepage}{S\arabic{page}}
\setcounter{figure}{0}\setcounter{table}{0}\setcounter{equation}{0}
\renewcommand{\theHfigure}{S\arabic{figure}}
\renewcommand{\theHtable}{S\arabic{table}}
\renewcommand{\theHequation}{S\arabic{equation}}
\captionsetup[figure]{name=Supplementary Fig.}
\captionsetup[table]{name=Supplementary Table}
\setlength{\parskip}{0.4ex}
\hypersetup{linkcolor=blue!45!black}

\begin{center}
{\bfseries\Large Supplementary Information\par}
\vspace{2mm}
{\large Continual Graph Memory and Community Infrastructure for AI-Assisted Mathematical Discovery\par}
\vspace{2mm}
{\small Jiang et al.\par}
\end{center}

\vspace{2mm}
\noindent\textbf{Contents}
\begin{enumerate}[leftmargin=2.2em,itemsep=0pt,topsep=2pt,label={}]
\item Supplementary Note 1. The \Ansatze\ platform: components, surfaces and public data (Supplementary Tables 1--3; Supplementary Figs.~\ref{sfig:hosted-runs}--\ref{sfig:sensitivity-lean})
\item Supplementary Note 2. Hosted CGM controller and analysis budgets (Supplementary Tables 4 and 5)
\item Supplementary Note 3. Formal specification of continual graph memory and the persistent runtime (Supplementary Tables 6 and 7; Supplementary Algorithms 1 and 2)
\item Supplementary Note 4. Manuscript analysis: verdicts and recorded cases (Supplementary Table 8; Supplementary Fig.~\ref{sfig:parity-report})
\item Supplementary Note 5. Formal verification states and recorded formal examples (Supplementary Table 9; Supplementary Fig.~\ref{sfig:even-snapshot})
\item Supplementary Note 6. Execution boundaries
\item Supplementary Note 7. Benchmark study: provenance, accounting and audits (Supplementary Tables 10--15)
\item Supplementary Note 8. Software verification and capture provenance (Supplementary Table 16)
\item Supplementary Note 9. Open problems: statements and claimed results
\item Supplementary Data and Software
\end{enumerate}
\noindent Values describe the inspected sources and archived records. Unless stated otherwise, platform values refer to source snapshot \code{b30740e} (13 October 2026); benchmark values are read from the archived records with no new model runs.

\section{Supplementary Note 1. The \Ansatze\ platform: components, surfaces and public data}
\subsection{Components and engines}
The platform comprises a public website (problem catalogue, reference atlas, news and quickstart), a console service that authorizes requests, creates jobs and serves artefacts, engine adapters that normalize heterogeneous runner output, persistent application state (SQLite for account-associated state; run directories for artefacts and graph caches) and a Lean checking interface reached through a fixed-interface broker. Registered engines can remain unavailable when a dependency, route or credential is missing; historical identifiers that are no longer active (for example \code{ucla} and \code{danus}) remain in old records as provenance and are not aliases for current implementations (Supplementary Table~1).

\stab{1}
\begin{table}[ht]
\centering\small
\caption{\textbf{Active engine identifiers at the inspected snapshot.} ``CLI registry'' is an integration category; it does not imply that every engine is installed or available to a member.}
\begin{tabularx}{\linewidth}{@{}L{3.2cm}L{2.4cm}Y@{}}
\toprule
Identifier & Registry class & Workflow represented\\
\midrule
\code{hermes} & Built-in & Embedded reasoning and tool harness\\
\code{improof} & Built-in & Author--critic ProofStack workflow\\
\code{math\_harness} & CLI registry & Hosted bounded CGM on an eligible account route\\
\code{kimi} & CLI registry & Hosted bounded CGM on the Kimi entry point\\
\code{codex}, \code{claude}, \code{openclaude}, \code{openhands}, \code{openclaw} & CLI registry & Coding-agent integrations\\
\code{deepagents} & CLI registry & DeepAgents integration with its own runtime\\
\code{plain} & CLI registry & Single-response baseline with saved proof output\\
\code{metaharness} & CLI registry & Solver, evaluator and proposer workflow\\
\code{deepseek\_harness} & CLI registry & DeepSeek-backed harness integration\\
\code{analyze} & CLI registry & Manuscript ingestion, review, roll-up and reporting\\
\bottomrule
\end{tabularx}
\end{table}

Member-facing proving choices are restricted to eligible tool-free routes; local tool-enabled harnesses and linked subscription capabilities require administrator policy. Existing runs retain their recorded execution configuration, so connecting another provider does not silently replace it.

\subsection{Surfaces and how to read them}
The workspace keeps one run selected while the researcher moves among its argument, activity, graphs and checks (Supplementary Table~2). A useful reading sequence begins with the problem and the proof, follows premises through Facts, consults Explore and Monitor when the path to the result matters, and then reads the formal target, assumptions, checking outcome and fidelity caveats. The next intervention may be editing prose, giving feedback, rechecking a manuscript result, formalizing or continuing an eligible attempt; repeating proof search is not interchangeable with these operations. Saving a revised proof uses an expected revision, so an unseen engine write produces a conflict; editing the argument does not establish that an earlier graph or Lean check supports the new content. A stale or missing result is informative: conflating it with a current negative verdict would discourage the right action, and conflating it with a current positive verdict would overstate the evidence.

\stab{2}
\begin{table}[ht]
\centering\footnotesize
\caption{\textbf{Interface surfaces, user actions and evidence boundaries.}}
\renewcommand{\arraystretch}{1.12}
\begin{tabularx}{\linewidth}{@{}L{2.3cm}L{3.2cm}L{4.2cm}Y@{}}
\toprule
Surface & Primary object & Researcher's action & Evidence boundary\\
\midrule
Composer & Problem or manuscript request & Select mode, source, model, route and context; submit & Launch settings are a request; the server records the effective configuration\\
Run list & Persisted execution records & Select, inspect, continue or delete an eligible run & A process state is not a verdict on the goal\\
Sandbox & Saved proof or Analyze report & Read, download, edit and give feedback where allowed & Edits and continuation may invalidate derived evidence\\
Monitor & Stage and agent events & Inspect a call and its available telemetry & Detail depends on what the adapter emitted; missing usage is not zero usage\\
DAG: Explore & Goals, plans, attempts, findings & Inspect alternatives, obstacles and route outcomes & Search relations are not mathematical dependencies\\
DAG: Facts & Accepted informal records & Follow premises and read the attached proof & Model acceptance is fallible and separate from Lean\\
DAG: Outline & Derived reading structure & Navigate a proof or manuscript & An outline is not an admission rule\\
DAG: Formal & Lean-derived declarations and links & Inspect types, provenance and relationships & Source links, compiler facts and fidelity judgements have different scopes\\
DAG: Manuscript & Segmented results and references & Filter verdicts, inspect blockers, recheck a result & Local judgement and dependency closure can disagree\\
Lean verification & Current Lean source and check metadata & Compile, Check, Verify or Audit when available & Read completion, axioms, source identity and audit state separately\\
Lean statements & Manuscript result signatures & Compile extracted statements & Proofs are not attempted\\
Library & Account-owned reusable context & Enable memory, skills, tools or context & Reuse does not certify truth or override runner policy\\
\bottomrule
\end{tabularx}
\end{table}

\subsection{Public discovery surfaces and data}
The Problems page supports search and filters for collection, topic, category, prize, AI involvement, Lean-related evidence and recent changes. A detail view links to the source formulation and to the evidence behind the status. Each problem has exactly one category derived from source-of-record markers, research records and explicit operator decisions; the current curation includes AI-assisted operator work from primary sources, so categories are not presented as independent sign-off by a human mathematician, and a live report can add a signal without becoming an accepted answer. Status overlays are tied to the catalogue content hash. The bundled catalogue artefact dated 11 October 2026 contains 3,084 problems from the Erd\H{o}s problems collection (1,221), the Kourovka notebook (1,307), a list of unsolved problems (464), a list of recently solved problems (39), Hilbert's problems (25), Smale's problems (18), the Millennium Prize problems (7) and Ramanujan's legacy (3)~\cite{si-ansatzeai}. These are snapshot counts; live status-category totals change with curation and are not reported as current measurements.

The atlas shows informal entries and formal declarations with distinct shapes, project colours, search and project and layer filters, and a list view. Unresolved or ambiguous references appear as placeholder endpoints; a detail panel keeps the original blueprint text or Lean type, resolution status, project provenance and pinned source links. Within-layer links keep the source's direction (a \code{USES} edge points from a citing entry to its reference; a \code{PROVES} edge from a proof environment to its statement), which differs from the premise-to-consumer convention of the per-run Facts view. These relationships are not forced into an acyclic graph, and no formal-to-formal dependency is inferred through a shared informal node. A retained \code{leanok} flag is source metadata, not a verification performed by the import. The Stacks-derived navigation graph~\cite{si-stacks} omits forward references to remain acyclic, so it is not the complete reference relation (Supplementary Table~3). News provides a cached feed of research signals with source links; a news item does not decide that a proof is correct.

\stab{3}
\begin{table}[ht]
\centering\small
\caption{\textbf{Sizes of bundled public reference artefacts at the inspected snapshot.} Counts are imported records and links, not distinct theorems or verified results. The catalogue artefact is dated 11 October 2026.}
\begin{tabularx}{\linewidth}{@{}Yrr@{}}
\toprule
Artefact & Records & Connections or scope\\
\midrule
Problem catalogue & 3,084 & eight source collections\\
Atlas: informal entries & 24,950 & 60 projects\\
Atlas: formal declarations & 16,682 & excludes placeholders\\
Atlas: unresolved endpoint placeholders & 3,892 & 3,895 unresolved or ambiguous links\\
Atlas: all correspondences & 20,656 & 16,761 resolved links\\
Atlas: informal dependencies & 54,200 & 54,128 uses; 72 proves\\
Stacks-derived reference DAG & 14,754 & 37,440 edges; 109 chapters; 75 forward and 485 self references omitted\\
\bottomrule
\end{tabularx}
\end{table}

\subsection{Reusable research memory and context admission}
The account library supplies owner-enabled context according to the selected engine's policy. The platform's published research memory is separate from the persistent harness's cross-project recall in Supplementary Note~3. Its records must originate from eligible completed runs, be authorized by the owner for public reuse and keep an atlas association and observed model provenance. Reusable formal records additionally require the current source to pass checking without placeholders, a faithful statement review without major mismatch and source-bound trusted checker evidence, conditions stronger than the ordinary \code{verified} display state. Platform context selection uses only eligible administrator-owned records, with at most four bounded excerpts; forum discussion is excluded. Retrieved content remains reference context and does not automatically become a locally admitted fact.

\begin{figure}[ht]
\centering
\includegraphics[width=\linewidth]{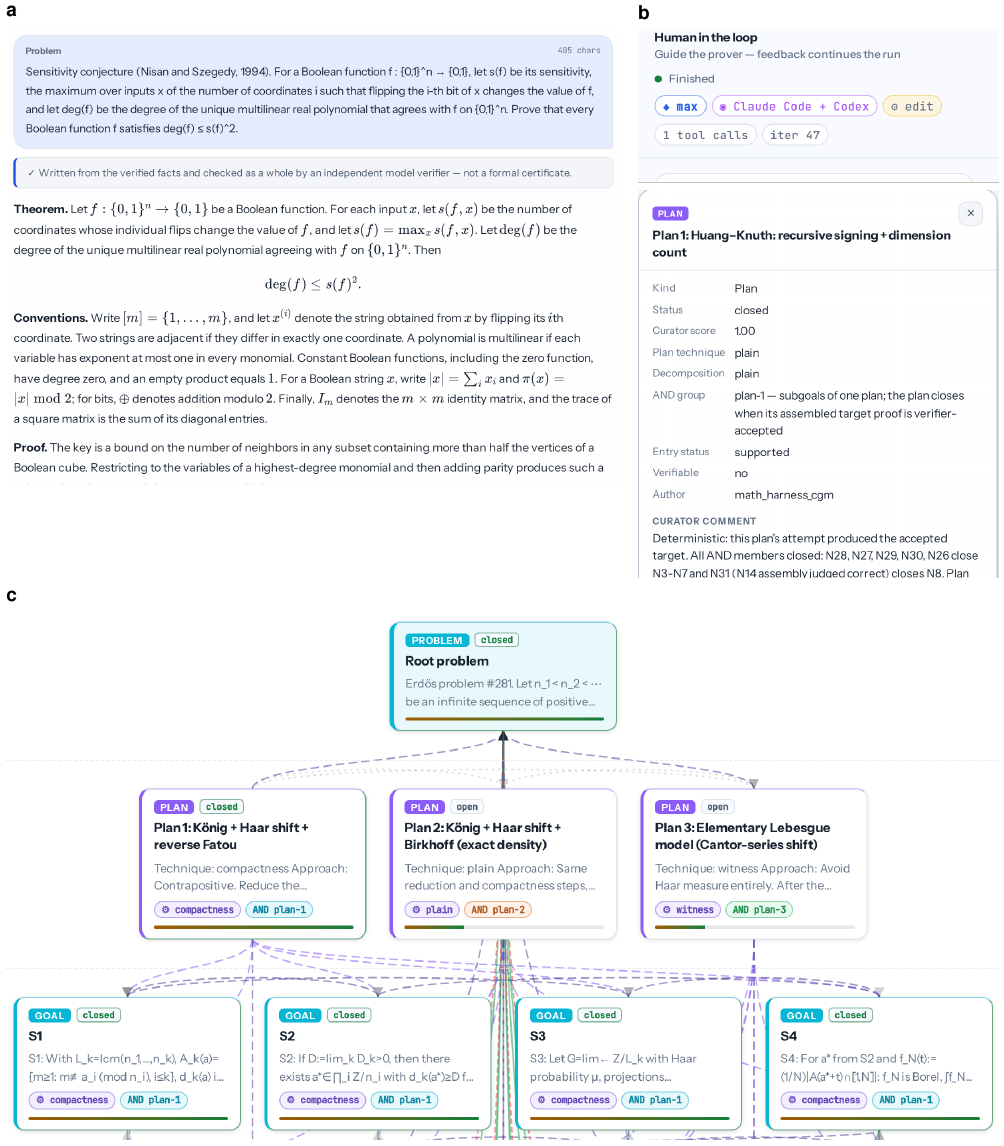}
\caption{\textbf{Hosted runs on the hybrid route in the workspace.} Captures of the console. \textbf{a},~Sandbox of the sensitivity run: the problem and the written proof composed from the verifier-accepted facts and checked as a whole by an independent model verifier, labelled as not a formal certificate. \textbf{b},~The run's route (\code{max}: Claude Code and Codex) and, from the Explore view, the curator's annotation of the closed plan (status, score, AND group and rationale). \textbf{c},~Explore view of the run on Erd\H{o}s problem 281, settled only in 2026: the strategist's three plans (K\"onig's lemma with a shift in the profinite completion and reverse Fatou, which closed; the same reduction with Birkhoff's theorem; an elementary Lebesgue model) and the closed subgoals of the first. The accepted targets are known theorems; the runs illustrate the display contracts.}
\label{sfig:hosted-runs}
\end{figure}

\begin{figure}[ht]
\centering
\includegraphics[width=\linewidth]{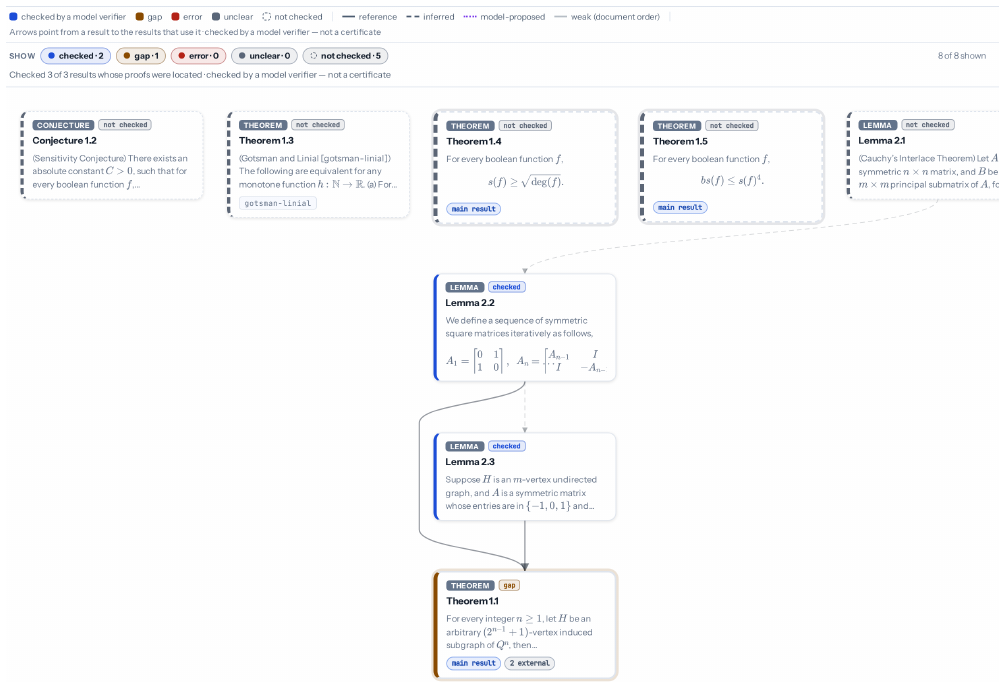}
\caption{\textbf{Manuscript graph of the recorded analysis of Huang's proof.} Verdict filters (2 checked, 1 gap, 0 error, 0 unclear, 5 not checked), the eight segmented statements and the dependency chain from Lemmas~2.2 and~2.3 to Theorem~1.1 taken from the manuscript's own references; dashed outlines mark statements without a located proof. Stored model judgements, not a certificate or a refereed review.}
\label{sfig:huang-analysis}
\end{figure}

\begin{figure}[ht]
\centering
\includegraphics[width=\linewidth]{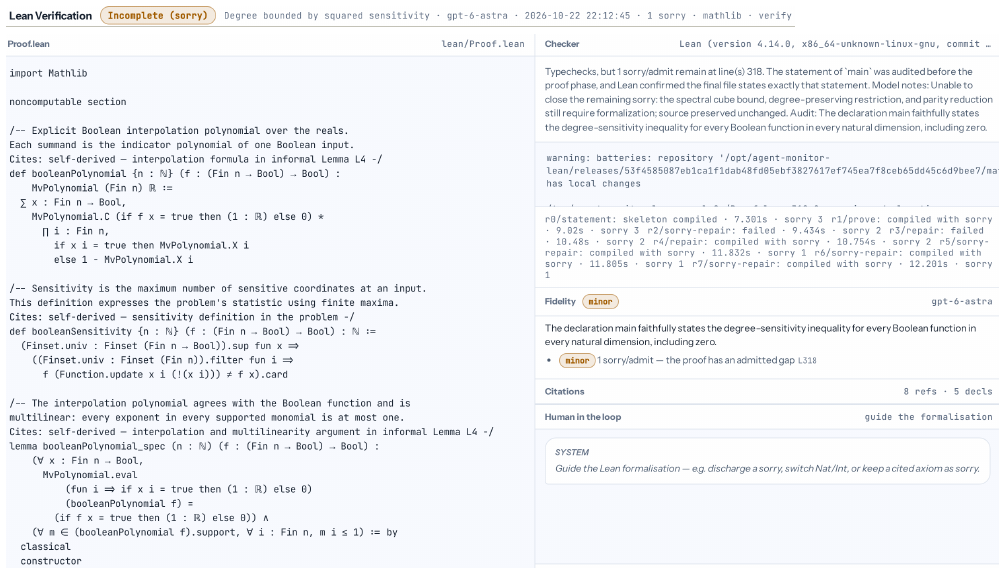}
\caption{\textbf{Lean verification of the sensitivity run.} The recorded state (\code{incomplete}, one placeholder), the start of the formalization (the interpolation polynomial and the sensitivity, with their provenance comments), the checker summary stating that the audited statement of \code{main} was preserved, the eight checking rounds and the fidelity audit. The Lean warning about local changes in a dependency is emitted by the pinned checking release and does not affect a verdict.}
\label{sfig:sensitivity-lean}
\end{figure}

\FloatBarrier

\section{Supplementary Note 2. Hosted CGM controller and analysis budgets}
\subsection{Stages, parsing and review}
Scaffold validates the workspace and problem without a model call. Explore invokes a strategist and the selected workers; each worker returns delimited lemma and theorem blocks, a self-reported outcome and optional findings. The parser checks unique labels, required statements and proofs, size limits, known dependencies, self-reference and acyclicity, and computes a deterministic topological order of the target's predecessor closure in which written order breaks ties; a forward reference is not itself a cycle. Ready blocks can be reviewed in waves, so concurrency need not preserve identical call consumption, but canonical candidate order rather than network arrival governs admission. The verifier sees accepted predecessor statements, not the search transcript or earlier proofs, through a fresh invocation; this separates interaction contexts without establishing statistically independent model errors. Critical errors or gaps override positive verdict text, substantive repair hints are incompatible with clean acceptance, and an unreadable response yields no acceptance. A stop before publication abandons pending work without a new result, a deadline can preserve settled material under the deadline policy, and abandoned replies cannot mutate saved artefacts. The receipt records the execution policy and call order without presenting response-arrival order as proof order.

\subsection{Budgets}
The hosted budget covers exploration, review, repair and curation (Supplementary Table~4). Recorded calls include model retries; the allowance is therefore not a count of independent proof attempts. Subscription transports default to one concurrent call. These hosted profiles apply to the platform controller and are separate from the persistent multi-worker benchmark configurations in Supplementary Note~7.

\stab{4}
\begin{table}[ht]
\centering\small
\caption{\textbf{Hosted CGM budget profiles.} The restricted profile is selected by the resolved access and funding policy; its defaults are also ceilings.}
\begin{tabularx}{\linewidth}{@{}Yrrr@{}}
\toprule
Control & Ordinary default & Ordinary ceiling & Restricted profile\\
\midrule
Recorded provider calls, including model retries & 16 & 64 & 12\\
Selected workers & 3 & 3 & 2\\
Repair attempts & 2 & 4 & 1\\
Candidates admitted to verification & 2 & 3 & 2\\
Lemma blocks per candidate & 8 & 16 & 6\\
Elapsed-time budget (s) & 5,400 & 14,400 & 2,400\\
Concurrent calls (API default) & 3 & 6 & 2\\
\bottomrule
\end{tabularx}
\end{table}

\stab{5}
\begin{table}[ht]
\centering\small
\caption{\textbf{Default Analyze caps by role and route.} A request can lower the applicable caps. Global ceilings are 256 result checks, 300 model calls and 14,400\,s. The restricted profile caps pasted text at 20,000 characters; other profiles permit up to 400,000.}
\begin{tabularx}{\linewidth}{@{}Yrrrr@{}}
\toprule
Profile & Results & Calls & Seconds & In flight\\
\midrule
Restricted access & 12 & 16 & 2,400 & 2\\
Member, sponsored & 30 & 36 & 3,600 & 2\\
Member, own key & 120 & 140 & 3,600 & 3\\
Administrator & 256 & 300 & 14,400 & 3\\
\bottomrule
\end{tabularx}
\end{table}
Long proofs may be checked in parts under the applicable prompt budget; the merged result keeps the caveat that reasoning across parts was not checked as one argument. Review coverage and the number of calls are different quantities.
\FloatBarrier

\section{Supplementary Note 3. Formal specification of continual graph memory and the persistent runtime}
\subsection{Research state and exploration-graph design}
For project $p_j$ at event $t$, the persistent state is
\begin{equation}
\mathcal M_{j,t}=\bigl(\{L_{i,t}\}_{i=1}^{m},\ C_t,\ F_t,\ E^F_t,\ E^X_t,\ A_t\bigr),
\end{equation}
where $L_i$ contains a worker's private notes, $C$ contains shared findings, $F$ contains accepted facts, $E^F$ records declared proof dependencies, $E^X$ records exploration relations and $A$ contains curator annotation events. The twelve shared-record types are plans, directions, conclusions, examples, counterexamples, proof attempts, obstacles, dead ends, strategic guidance, progress syntheses, verification records and reports. Membership in $C$ does not imply admission to $F$. Exploration nodes reference these records, accepted facts or symbolic goals of the form \code{goal:<slug>}; the graph does not duplicate their mathematical content.

Workers record relations when publishing findings. The fourteen edge types are \code{addresses}, \code{decomposes-into}, \code{alternative-of}, \code{depends-on}, \code{supports}, \code{refutes}, \code{blocks}, \code{refines}, \code{promoted-to}, \code{supersedes}, \code{prioritizes}, \code{reduces-to}, \code{equivalent-to} and \code{generalizes}. Each edge records its endpoints, type, author, timestamp and rationale. Decomposition edges can share an \code{and\_group} and carry a technique tag: \code{plain}, \code{localization}, \code{invariant}, \code{extremal}, \code{compactness} or \code{witness}. The group identifies jointly required obligations, including reassembly for a localization plan. Alternative plans remain separate routes. Shared records, relation events and annotation events are append-only; adjacency and current views are derived when requested.

An active role $r$ chooses actions from its contract, private history $h^r_{j,t}$ and retrieved memory $x^r_{j,t}$:
\begin{equation}
a^r_{j,t}\sim\pi^r_\theta\bigl(\cdot\mid g_j,h^r_{j,t},x^r_{j,t},\mathrm{Contract}^r\bigr),\qquad (\mathcal M_{j,t+1},o_{j,t+1})=\mathrm{Execute}(\mathcal M_{j,t},a^r_{j,t}),
\end{equation}
where $g_j$ is the project target and $o$ is the returned observation. Model parameters remain fixed within a deployment. Workers and the curator operate asynchronously; Supplementary Algorithms~1 and 2 summarize their interactions rather than impose a synchronized schedule.

\subsection{Fact admission, promotion and revocation}
A candidate records the problem $p$, statement $S$, proof $\Pi$, declared predecessors $P$, glossary $G$ and bibliographic metadata. Its identifier is the first 16 hexadecimal characters of $\mathrm{SHA256}(\mathrm{canon}(p,P,G,S,\Pi))$. Canonicalization sorts predecessors and glossary entries and normalizes statement and proof whitespace. Bibliographic metadata is excluded, allowing citation corrections without changing dependency identifiers.

Admission requires deterministic prechecks, an accepting verdict from a fresh verifier invocation and a completed write. Revoked predecessors are checked before verification and again at persistence. For a successfully written fact $u$, admission adds $u$ to $F_t$ and $(r,u)$ to $E^F_t$ for every declared predecessor $r\in P$. Revocation removes the selected fact and its declared descendants from the active store:
\begin{equation}
R_t(u)=\{u\}\cup\mathrm{Desc}_{E^F_t}(u),\qquad F_{t+1}=F_t\setminus R_t(u).
\end{equation}
Once a verdict exists, the gateway records it together with the report, returned identifier and any write failure in shared memory. An accepting verdict without a successful write therefore does not create a fact. If the worker supplies a valid originating finding identifier, the gateway also records its fact identifier, advances its status to \code{verified} for a verifiable finding or \code{supported} for a judgment, and adds a \code{promoted-to} edge unless that edge already exists. This back-link is advisory: failure to record it does not reverse an already completed fact write.

\begin{proposition}[Local admission property]
Starting from an empty fact store, if every addition uses the default gateway and persistence succeeds, every admitted fact has an accepting verdict from a separate verification invocation, and changing curator scores alone cannot admit a fact.
\end{proposition}
\noindent This property follows from the gateway's write condition and role-specific tool exposure. It describes the admission mechanism, not verifier soundness. The persistent store rejects revoked predecessors but does not itself enforce predecessor existence or acyclicity; those requirements remain part of the agent workflow, unlike the structural checks of the hosted controller in Supplementary Note~2.

\subsection{Curator annotations, propagation and frontier ranking}
An annotation $a=(v,z,s,e,c,\tau)$ records a node, status, optional score $s\in[0,1]$, evidence identifiers, explanatory comment and timestamp. Status is \code{open}, \code{active}, \code{blocked}, \code{dead} or \code{closed}. The last appended annotation supplies the current view, denoted $\widehat A_t(v)$; an absent status defaults to \code{open}. Earlier events remain available for inspection. The curator contract requires judgments to cite evidence and directs the agent to fetch complete findings and verification reports before assessing them.

The curator inspects batches of changes after verification results, accumulated findings or new strategic guidance. Its propagation rules follow the recorded argument structure. An AND route requires every obligation, including reassembly where applicable; failure of one obligation invalidates that route without invalidating alternative routes. A reduction from $P$ to $Q$ requires both completion of $Q$ and an accepted bridge $Q\Rightarrow P$. Equivalence requires facts supporting both directions. A proved generalization supports its special case through a specialization fact, whereas a refuted special case refutes the generalization. Refuting or blocking evidence must be verified or otherwise decisive before a route is marked dead or blocked. These are instructions interpreted by the curator; the storage layer validates types and score ranges without performing mathematical inference.

A node is marked stale when a newer relation touches it:
\begin{equation}
\mathrm{stale}_t(v)=\mathbf 1\!\left[\max_{e\ni v}\tau(e)>\tau(\widehat A_t(v))\right].
\end{equation}
Missing timestamps take $-\infty$. This check uses incident-edge timestamps, not semantic changes to the records named in the evidence list. The frontier candidate set $\mathcal V_t$ contains open or active plans, directions and annotated symbolic goals. With $d^X_t(v)$ incident edges, an unset score represented by $\bot$, and $\bar s_t(v)=0$ for an unset score, its ordering is
\begin{equation}
\kappa_t(v)=\bigl(\mathbf 1[d^X_t(v)=0],\ \mathbf 1[s_t(v)=\bot],\ -\bar s_t(v),\ c_t(v)\bigr),\qquad \mathrm{Frontier}_K(t)=\mathrm{first}_K\,\mathrm{sort}_{\mathrm{lex}}(\mathcal V_t;\kappa_t).
\end{equation}
Here $c_t(v)$ is the finding's creation timestamp, or the current annotation timestamp for a symbolic goal. Linked nodes precede unlinked nodes, scored nodes precede unscored nodes within each group, and higher scores rank first. The curator is instructed to leave unlinked proposals unscored until workers connect them. Scores combine estimated value for the target with a heuristic exploration bonus; immediate dependent counts inform the assessment of prerequisites. Strategic guidance overrides local rankings. Neither a high score nor a \code{closed} exploration status admits a fact.

\subsection{Graph-based retrieval}
Lexical search uses BM25 (ref.~\citenum{si-bm25}) over shared records or fact documents:
\begin{equation}
\mathrm{BM25}(q,d)=\sum_{x\in\mathrm{uniq}(q)}n(x,q)\,\mathrm{idf}(x)\frac{(k_1+1)\,n(x,d)}{n(x,d)+k_1(1-b+b|d|/\overline L)},\qquad \mathrm{idf}(x)=\log\Bigl(1+\frac{N-\mathrm{df}(x)+\tfrac12}{\mathrm{df}(x)+\tfrac12}\Bigr).
\end{equation}
Here $N$ is corpus size, $n(x,d)$ is term frequency, $\mathrm{df}(x)$ is document frequency and $\overline L$ is mean document length. The implementation uses $k_1=1.5$ and $b=0.75$, with lowercased ASCII letters, digits and underscores as tokens. Empty queries or corpora score zero. Shared-finding search ranks full records but returns metadata and bounded claim previews; selected records are then fetched in full, including evidence, reports and current status. Fact search returns identifiers and statements, with proofs read separately from the fact records. Search views are rebuilt from stored content on demand.

For a selected fact $u$, neighbourhood retrieval first returns declared predecessors in stored order. Optional immediate dependents consume the next portion of the requested budget, and their count is always returned. The remaining portion is used for accepted facts linked directly to $u$ by an exploration edge, or through two exploration edges whose intermediate node is not an accepted fact. Traversal considers both directions, while paths retain the original relation labels and directions. Exploration candidates follow edge-log order and are deduplicated; candidates already returned as predecessors are removed after candidate selection, so the result can contain fewer items than the budget allows. These operations return statements and paths rather than full proofs. Exploration proximity supplies context without adding proof dependencies.

\subsection{Cross-project experience and curator lessons}
Cross-project access is disabled by default. When enabled, it covers projects with the same batch label or all other accessible projects. The filter enforces membership; chronological ordering and the availability of completed projects are controlled by the run protocol. Workers and the main agent can request accepted fact statements, negative findings (dead ends, obstacles and counterexamples), or curator lessons. Strategic guidance and progress syntheses are excluded.

Each channel is ranked independently. For requested channels $\mathcal C$ and eligible corpus $\mathcal D_c$,
\begin{equation}
\mathcal R_c(q)=\mathrm{Top}^{+}_{k}\{(d,\mathrm{BM25}(q,d;\mathcal D_c)):d\in\mathcal D_c\},\qquad k=\max\bigl(1,\lfloor \mathrm{limit}/|\mathcal C|\rfloor\bigr),
\end{equation}
where $\mathrm{Top}^{+}$ retains only positive-scoring results. Scores are channel-specific and are not jointly normalized. Returned items retain source-project provenance: fact results contain statements and identifiers, negative findings contain claims and short evidence excerpts, and lessons contain patterns and proposed adjustments. A recalled mathematical statement requires proof and admission in the receiving project before use as a local premise. Under batch filtering, eligible lessons can include those from the current project as well as other projects with the same label.

A lesson $\ell=(\mathrm{pattern},\mathrm{evidence},\mathrm{adjustment})$ records a scoring habit, the observations motivating it and a concrete change for subsequent curation. Each submitted field must be non-empty after whitespace stripping and no longer than 2,000 characters before stripping; the stored fields are stripped. The curator contract restricts lessons to methodology, while the code checks field presence and length. At project close, the curator can use its annotation history and recorded route outcomes to formulate a retrospective. A new curator receives its standing contract followed by the latest 20 eligible lessons when cross-project memory is enabled. This changes the context available to the agent without changing model parameters or the standing contract. Retrospectives are optional and may be absent after forced termination.

\begin{algorithm}[h]
\caption{Continual search with persistent graph memory}
\small
\begin{algorithmic}[1]
\Require projects $(p_j,g_j)_{j=1}^{J}$, budgets, memory mode $\sigma$; operator-controlled visibility and sequencing
\For{$j=1,\dots,J$}
  \State establish eligible scope $\mathcal U_\sigma(p_j)$ and initialize the project stores
  \State launch the main agent, workers and curator; supply the curator with eligible recent lessons
  \While{the main agent continues the run}
    \State workers read assignments and retained state, explore, and append findings and typed relations
    \State answer memory requests with \textsc{ReadMemory}; process submitted proofs with \textsc{Admit}
    \State on a curator wake, read changed evidence, apply propagation rules and update annotations
    \State the main agent adjusts assignments, consultation and stopping
  \EndWhile
  \State at orderly close, optionally record retrospective lessons
  \State under a sequential protocol, stop remaining workers before exposing the completed project to later runs
\EndFor
\end{algorithmic}
\end{algorithm}

\begin{algorithm}[h]
\caption{Memory access and admission primitives}
\small
\begin{algorithmic}[1]
\Procedure{ReadMemory}{$q,u,p,\sigma$}
  \State search project-local findings and facts; fetch selected findings in full when their evidence is needed
  \State if $u$ is a selected fact, retrieve its predecessors and requested graph neighbours with statements and paths
  \State if cross-project recall is requested and enabled, return channel-ranked results with provenance
\EndProcedure
\Procedure{Admit}{$f$, source\_id}
  \State initialize $u\gets\bot$; check revoked predecessors and deterministic preconditions
  \State on a precheck failure, return error and repair information
  \State obtain a verdict and report from a fresh verifier invocation; on operational failure, return an error
  \State if the verdict is \code{correct}, attempt persistence after rechecking revoked predecessors
  \State set $u\gets\mathrm{id}(f)$ only if the write succeeds
  \State append the verdict, report, $u$ and any write error to shared memory
  \State if $u\neq\bot$ and source\_id resolves, update the originating finding and add its advisory promotion link
  \State return the verdict, identifier and repair or error information
\EndProcedure
\end{algorithmic}
\end{algorithm}

\subsection{Roles, lifecycle and services}
\stab{6}
\begin{table}[ht]
\centering\small
\caption{\textbf{Ordinary gateway tool visibility by role.} Thirteen gateway operations are grouped into eleven capabilities where access is identical. Unknown roles fall back to the verifier set; an explicit development role can expose all tools. These are gateway permissions, not filesystem isolation.}
\begin{tabularx}{\linewidth}{@{}Ycccc@{}}
\toprule
Capability & Main & Worker & Verifier & Curator\\
\midrule
Record a research finding & yes & yes & -- & --\\
Search and read research memory & yes & yes & -- & yes\\
Submit a candidate fact & -- & yes & -- & --\\
Search facts and inspect dependencies & yes & yes & -- & yes\\
Revoke an accepted fact & yes & -- & -- & --\\
Search published theorems & yes & yes & yes & --\\
Link research records & -- & yes & -- & --\\
Annotate the exploration graph & -- & -- & -- & yes\\
Inspect the research frontier & yes & yes & -- & yes\\
Record a methodological lesson & -- & -- & -- & yes\\
Recall earlier research & yes & yes & -- & --\\
\bottomrule
\end{tabularx}
\end{table}

\stab{7}
\begin{table}[ht]
\centering\small
\caption{\textbf{Persistent lifecycle commands.}}
\begin{tabularx}{\linewidth}{@{}L{2.8cm}Y@{}}
\toprule
Verb & Effect and boundary\\
\midrule
\code{new} & Create the project and requested worker roles; workers remain unstarted\\
\code{assign} & Replace one worker's task file; an assignment update, not a fact submission\\
\code{start} & Launch detached worker loops with a process lock against duplicate starts\\
\code{status}, \code{list} & Read liveness, round progress, recent outcomes and project summaries\\
\code{stop} & Request a graceful boundary stop; an explicit force option can terminate the worker process group\\
\code{finalize} & Record selected target identifiers, or suggest terminal facts; does not stop workers\\
\bottomrule
\end{tabularx}
\end{table}
A worker round reads its assignment and recovers context from the persistent stores. Restarting a worker launches a new model session; accepted facts, shared findings and private notes remain available. The loop respects stop requests, deadlines, round limits and operational failure limits. Completed bounded assignments can be parked until reassignment, and repeated idle rounds use a backoff delay. Verification failures caused by missing output, malformed responses or timeouts do not admit facts. Strategy consultation supplies guidance and records its usage separately from fact admission. These lifecycle controls support continuation across sessions without treating process activity as mathematical progress.

\noindent\textbf{Scope of enforcement.} Role-specific gateway tools separate fact submission, revocation, relation writing and curator annotation. They do not provide filesystem isolation. Endpoint checks for exploration edges are advisory; symbolic goals need no separate stored record. Curator propagation and scoring follow the operating contract, while the code constructs the frontier and flags newer incident edges. Cross-project ordering is determined by the run protocol, and lesson validation checks fields rather than mathematical content.

\section{Supplementary Note 4. Manuscript analysis: verdicts and recorded cases}
\stab{8}
\begin{table}[ht]
\centering\small
\caption{\textbf{Analyze verdicts describe model review of a submitted result.} None is a formal certificate.}
\begin{tabularx}{\linewidth}{@{}L{2.4cm}Y@{}}
\toprule
Verdict & Interpretation and next inspection step\\
\midrule
\code{checked} & The parsed review recorded no critical error or gap; inspect the argument, assumptions and dependency scope\\
\code{gap} & A missing justification or unresolved step; inspect the cited location and supporting results\\
\code{error} & A critical error; inspect the specific claim before accepting the diagnosis\\
\code{unclear} & An interpretable review without a clear assessment; inspect its stated reason\\
\code{not\_checked} & No usable judgement (budget, source length, interruption or response format)\\
\bottomrule
\end{tabularx}
\end{table}
An uninterpretable reply may be stored with raw verdict \code{unclear} and \code{interpreted=false}; consumers must keep the flag, because an interpretable uncertain judgement differs from no judgement. Definitions, assumptions and external references can transmit predecessor issues without acquiring an invented proof judgement.

\textit{Illustrative fixture.} The Analyze captures use an original four-result parity manuscript with scripted reviews. Its first lemma correctly states that the sum of two even integers is even; its second lemma incorrectly states that the product of two odd integers is even; a theorem derives that an odd square is even from the false lemma; and a corollary repeats the implication for positive odd integers. The reviews mark the first lemma checked, the second erroneous and the theorem locally checked relative to its premise, and a reduced budget leaves the corollary unchecked; the roll-up exposes the upstream error in both consumers (Supplementary Fig.~\ref{sfig:parity-report}). The fixture demonstrates the contract, not a model's ability to detect the error.

\textit{Recorded analysis.} A user-selected analysis concerns a didactic manuscript about finite-time blow-up in the smoothly forced three-dimensional Navier--Stokes equations. Its saved process state is \code{finished}. It reports reviews of all 70 results with located proofs, while its 81 statement nodes include 10 \code{checked}, 60 \code{gap} and 11 \code{not\_checked} entries; Theorem~1.1 has a local gap and a dependency-closure gap, and Theorem~3.1 is not checked because the parser did not locate its proof. No Lean artefact is present. Its title and attribution remain claims of the source document, and the verdicts are stored model judgements about that manuscript, not established results of this paper or an independent mathematical review. The stored content was replayed in the read-only viewer at commit \code{c37224c} without a new model or checker call.

\begin{figure}[ht]
\centering
\includegraphics[width=0.86\linewidth]{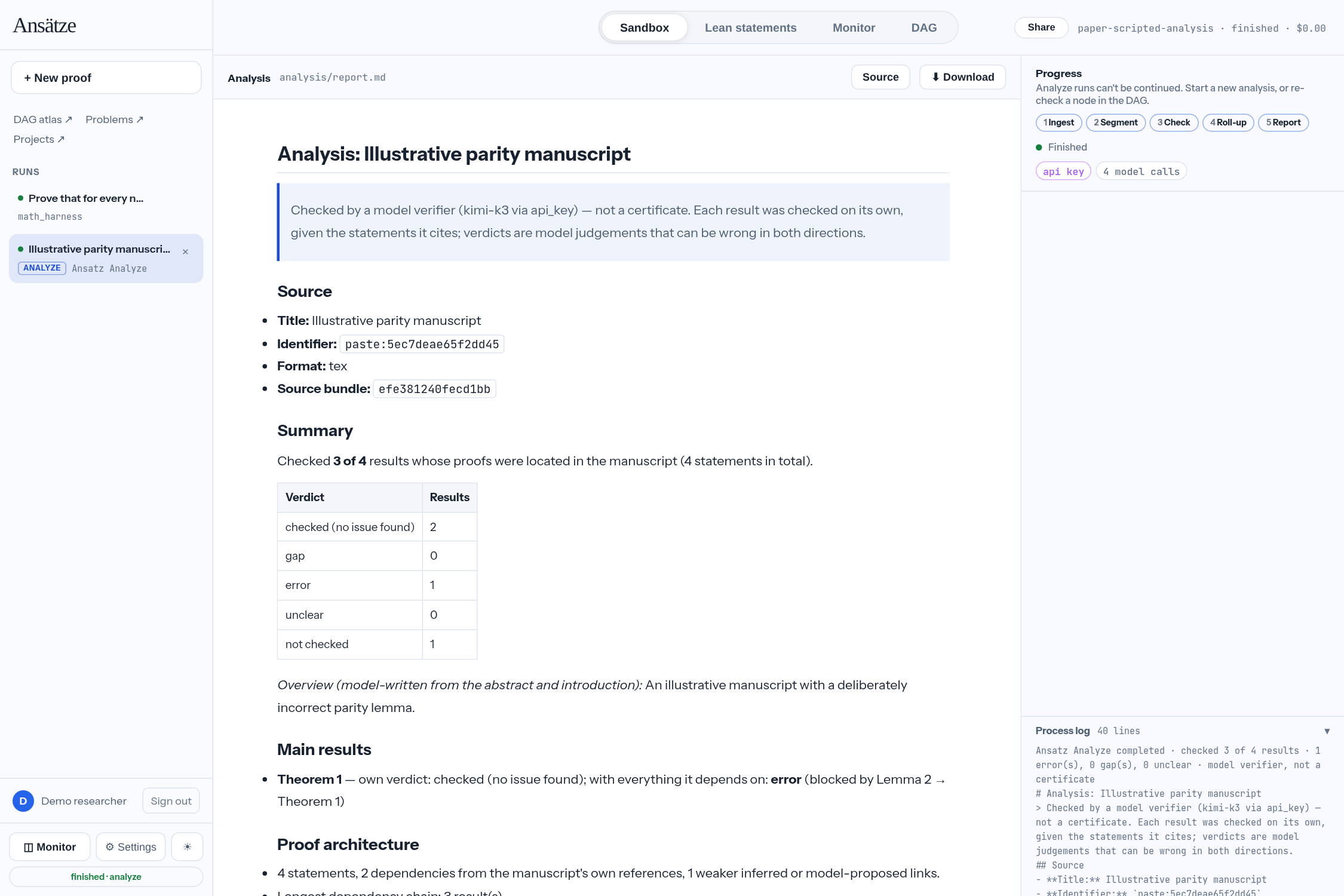}
\caption{\textbf{Analyze report for the illustrative parity manuscript.} The report states that 3 of 4 results with located proofs were checked by a model verifier, lists the verdict counts (2 checked, 1 error, 1 not checked) and records that Theorem~1 is locally checked but blocked through Lemma~2. Scripted reviews; not a measurement of model accuracy.}
\label{sfig:parity-report}
\end{figure}
\FloatBarrier

\section{Supplementary Note 5. Formal verification states and recorded formal examples}
\stab{9}
\begin{table}[ht]
\centering\small
\caption{\textbf{Formal result states and their limited interpretation.} Fidelity metadata must be read separately, especially when an audit is unavailable or the source was edited.}
\begin{tabularx}{\linewidth}{@{}L{3.4cm}Y@{}}
\toprule
State & What the recorded action establishes\\
\midrule
\code{compiled} & Compilation succeeded; placeholders may remain\\
\code{incomplete} & The file elaborated, but completion checks found unresolved placeholders\\
\code{verified} & The checked source passed completion and soundness conditions with no major mismatch detected by the available fidelity checks; does not by itself imply a successful model audit\\
\code{unfaithful} & A major statement mismatch was found for an otherwise checked or incomplete candidate\\
\code{failed} & The requested checking action failed or the submitted source was refused\\
\code{toolchain\_missing} & The authoritative checker could not be used; no successful check is inferred\\
\bottomrule
\end{tabularx}
\end{table}
Several combinations that may look contradictory are valid: a run can be \code{finished} with an unaccepted candidate; a CGM target can be model-accepted while its Lean file is \code{incomplete}; a Lean file can be \code{compiled} with placeholders or kernel-checked while its fidelity audit is unavailable; an Analyze result can be locally \code{checked} yet inherit an \code{error}; and a formal result can become stale after the proof changes. Compile and Check operate on current source without a new model call, so a manually supplied artefact can pass formal checking without a new fidelity audit, and an audit-provider failure does not invalidate an otherwise valid Lean proof.

\textit{Public odd-integer artefact.} The public quickstart artefact asks whether the sum of two odd integers is even. Captured on 27 September 2026, it contains the argument, four admitted informal facts with five dependency edges, an exploration graph of 33 nodes and 51 edges that retains the successful plan and a dead alternative, Lean source and a recorded checking result. Its target preserves the integer domain and arbitrary inputs:
\begin{lstlisting}
theorem main (a b : ℤ) (ha : Odd a) (hb : Odd b) : Even (a + b) := by
  exact odd_add_odd_even a b ha hb
\end{lstlisting}
Each informal fact is linked to a named declaration; this is complete mapping coverage for the four records, not an independent fidelity audit of every helper. The independent recheck is described in Supplementary Note~8.

\textit{Even-integer snapshot.} A user-selected snapshot proves that the sum of two even integers is even with two model-accepted lemmas and a target (three saved \code{correct} verdicts; informal and formal graphs of three nodes and two edges). It includes saved Lean source but no checker result or fidelity verdict, and the interface displays both absences (Supplementary Fig.~\ref{sfig:even-snapshot}). The conclusion supported by inspection is that a formal artefact is available for checking, not that a check succeeded; the older export also lacks separate Explore and Facts artefacts, and their absence is a property of the saved material.

\begin{figure}[ht]
\centering
\includegraphics[width=0.49\linewidth]{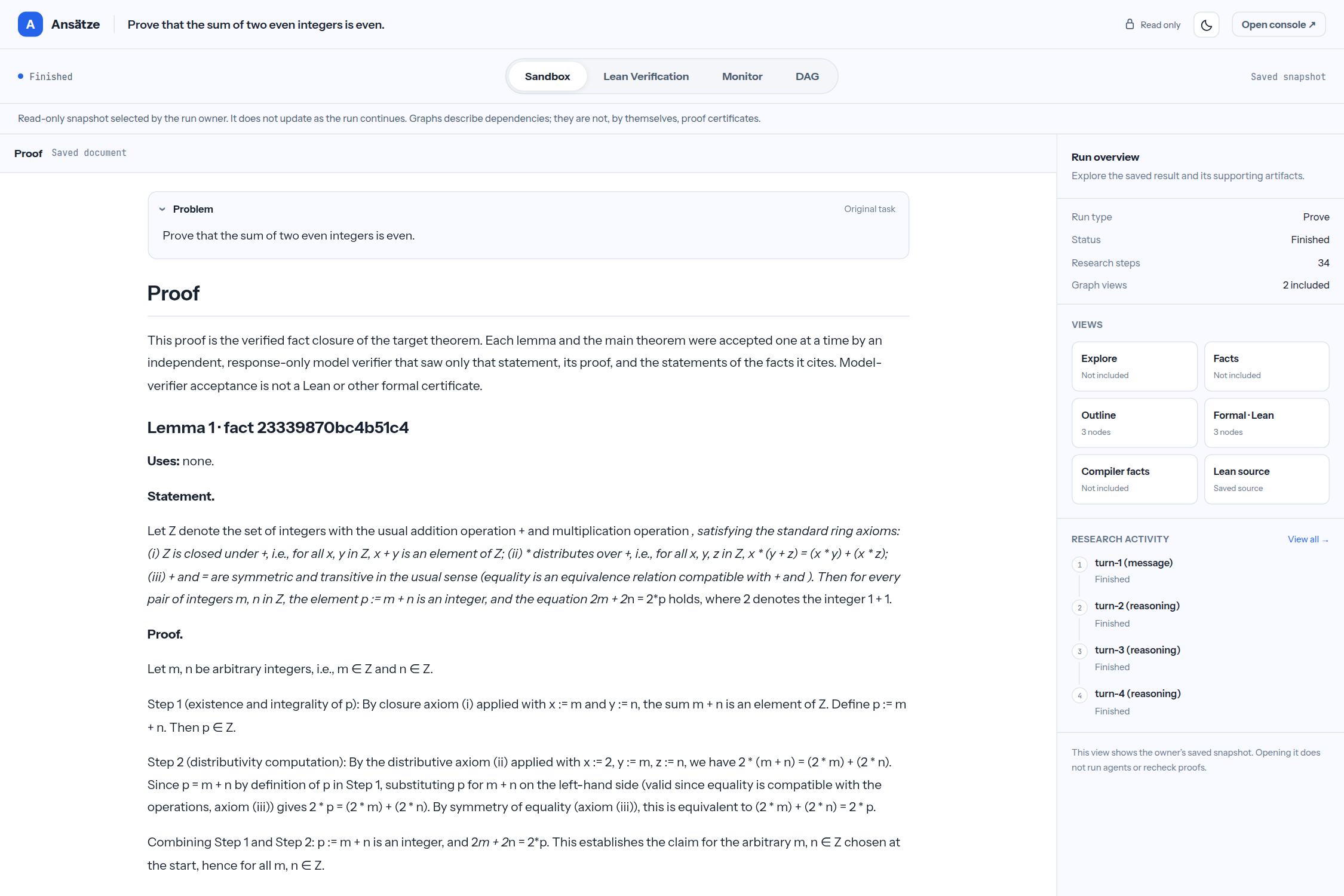}\hfill
\includegraphics[width=0.49\linewidth]{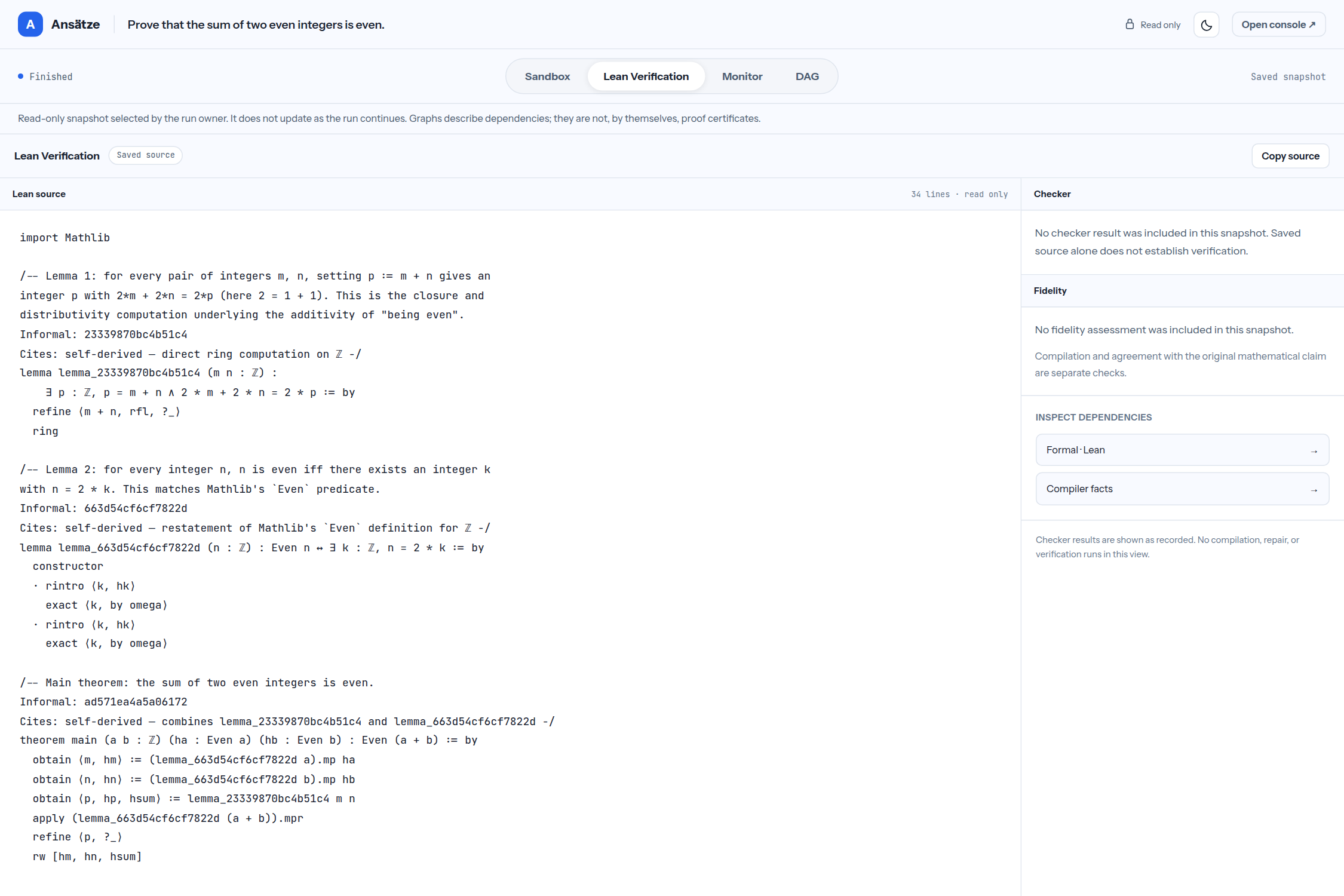}
\caption{\textbf{A recorded snapshot whose formal check is missing.} Left, the saved informal argument for the even-integer claim, with two cited premises. Right, Lean verification of the same snapshot: the saved helper declarations and target are shown, with separate notices that no checker result and no fidelity assessment were included. Plausible Lean source is not presented as compilation or verification.}
\label{sfig:even-snapshot}
\end{figure}
\FloatBarrier

\section{Supplementary Note 6. Execution boundaries}
The platform and persistent harness enforce different execution contracts. Hosted models return text for controller parsing, budget checks, review and artefact construction. Persistent agents use role-specific memory tools within the host environment; a fresh working directory does not itself provide process isolation. Manuscript ingestion checks paths, file types, sizes and network destinations before extracting content. Formal-check records bind a result to its source and toolchain under the trusted checking host. These boundaries explain the provenance and checking states described in the main text; they do not establish mathematical correctness by themselves.

\section{Supplementary Note 7. Benchmark study: provenance, accounting and audits}
\subsection{Data sources and figure correspondence}
The benchmark analysis uses archived accounting exports, normalized records, operator ledgers and summaries, and post-hoc review files (Supplementary Data~1 and 2). In the Second Batch figure, panel a presents the reported system comparison, panel b the complete ten-task ablation outcomes, and panel c the fresh-verifier re-review. Panels d and e aggregate the recorded tokens and fact snapshots for Ansatz, and panel f reports its neighbourhood-retrieval and cross-project-recall calls. The Ansatz reference in panel b uses the study's grading criterion. Missing proof chains, transcripts or configuration logs are not reconstructed from summary counts.

\subsection{External results}
The local problem statements reproduce the ten tasks in the First Proof Second Batch report~\cite{si-firstproof2}. Supplementary Table~10 gives the published editorial decisions and referee recommendations for the four official teams shown in main-text Fig.~\ref{fig:audit}a. A pass is an essentially flawless submission or one requiring minor revisions; major revisions and rejections do not pass. ETH submitted no result for task 06, which remains in its denominator. The reported Plain, Codex, Danus and Ansatz outcomes in Fig.~\ref{fig:audit}a use the author-supplied comparison data; their respective totals are 4, 6, 7 and 10.

\stab{10}
\begin{table}[ht]
\centering\scriptsize
\caption{\textbf{First Proof Second Batch: published editorial decisions and referee recommendations.} Columns A--D correspond to ETH, UCLA, OpenAI and Princeton in main-text Fig.~\ref{fig:audit}a, respectively. Decisions: F, essentially flawless; m, minor revisions; M, major revisions; R, reject. Referee recommendations: mC, minor (citation); mM, minor (mathematical correction).}
\renewcommand{\arraystretch}{1.12}
\begin{tabularx}{\linewidth}{@{}L{2.4cm}L{2.6cm}cccc*{4}{>{\raggedright\arraybackslash}X}@{}}
\toprule
Task & Domain & A & B & C & D & Referees A & Referees B & Referees C & Referees D\\
\midrule
01 Automorphism & Computability theory & m & m & m & R & F,mC & F,mC,mC & F,mC & R,R\\
02 M\"obius & Discrete geometry & m & m & m & R & mC,mC & mM,mC,mC & mM,mM & R,R,R\\
03 Bernoulli & Discrete probability & m & R & R & R & mC,mC,mC & R,R,R & R,R & R,R\\
04 Dilation & Metric geometry & R & R & R & R & M,R & R,R,R & R,R & R,R,R\\
05 SPDE & Stochastic PDE & F & R & R & R & F,F,F & R,R,R & R,R,R & R,R\\
06 Tree lattice & Lattice theory & -- & F & F & R & -- & F,F,F & F,F,F & R,R,R\\
07 Matching & Combinatorial topology & F & m & F & m & F,F,mM & mC,mM,M & F,mC,mM & F,mC,mM\\
08 Dressian & Matroids and tropical geometry & R & M & R & R & M,M,R & M,M & R,R,R & R,R\\
09 Hilbert & Algebraic combinatorics & m & m & m & R & mC,mM & mC,mM,mM & mM,mC & R,R,R\\
10 von Neumann & von Neumann algebras & M & R & M & R & M,M & R,R & M,M,M & R,R\\
\midrule
Passes (F or m) & & 6 & 5 & 5 & 1 & & & & \\
\bottomrule
\end{tabularx}
\end{table}

\subsection{Second Batch: evaluation protocol and accounting conventions}
The organizers evaluated the four external systems under a 24-hour limit, with no human interaction after input and two or three referees per submission~\cite{si-firstproof2}. The reported Ansatz total follows the study's grading criterion: nine full solutions plus the partial Bernoulli characterization, which also received a pass. Thus, ten passes do not mean ten complete target proofs. Phase-level outcomes in the accounting records refer to whether the full target was solved under the harness's review procedure. Separate grading sessions compare recorded conclusions with published answers and referee summaries; answer-containing materials are excluded from solving sessions. Published expert decisions and the harness's model-based assessments retain their respective evaluation procedures.

Ansatz ran in an isolated deployment for the Second Batch, with web search and fetching disabled for all roles. Strategy consultation used the worker/verifier model at maximum effort without tools. The initial phase solved eight full targets; continuation of Dilation added the ninth, while Bernoulli remained partial. The two continuations retained their project state. A forced architecture consultation before the Dilation continuation is recorded as an operator intervention. Main-text Fig.~\ref{fig:audit}d,e combines the recorded phases by project rather than treating continuations as independent problems.

An accounting window spans from five minutes before the first worker start to ten minutes after the last round-log modification unless forced-stop bounds were supplied; it is not time to proof. Worker tokens sum printed token footers or, for killed sessions, recovered totals minus cached input. Verifier tokens are attributed by time window. Main-agent and curator counts are new tokens (input, output and cache writes, excluding cache reads); consultation charges are strategy-API spending only. Across the twelve accounting exports for Ansatz, 27 worker rounds and six verifier runs lack token lines, so totals are recorded rather than complete usage.

\subsection{Ansatz configuration}
\stab{11}
\begin{table}[ht]
\centering\footnotesize
\caption{\textbf{Ansatz configuration for First Proof Second Batch.} The ablation configurations are described separately below.}
\renewcommand{\arraystretch}{1.12}
\begin{tabularx}{\linewidth}{@{}L{4.2cm}Y@{}}
\toprule
Setting & Ansatz\\
\midrule
Dates (2026) & 13--17 September\\
Workers and roster & 10 (\code{high:5,xhigh:4,max:1})\\
Worker and verifier model & GPT-5.6 Sol\\
Main agent and curator & Claude Code sessions\\
Memory policy & Fact graph and curated exploration graph; neighbourhood retrieval; batch-scoped recall of accepted fact statements, negative findings and curator lessons\\
Web access & Web search and fetch denied for every role\\
Caps & 360 min first pass + 40 min backstop; uncapped continuations for tasks 03 and 04\\
Intervention & Forced architecture consultation before solving task 04\\
Full targets solved & 8/10 first pass; 9/10 after continuation\\
Study-criterion passes & 10/10, including the Bernoulli partial result\\
\bottomrule
\end{tabularx}
\end{table}

\subsection{Ansatz: accounting}
Supplementary Table~12 lists the twelve accounting exports underlying main-text Fig.~\ref{fig:audit}d,e. Summing recorded tokens across the initial and continuation phases gives 922,003,391 tokens. Fact counts use only the latest snapshot of each project, giving 5,013 accepted facts. Bernoulli and Dilation contribute 322.0 and 281.0 million tokens and 4,102 and 586 facts, respectively. Accounting windows and role-specific log intervals are not identical: in particular, the resumed Bernoulli main-agent and curator sessions extend beyond its worker window. Tokens describe recorded activity rather than a uniformly timed proof-search interval.

\stab{12}
\begin{table}[ht]
\centering\scriptsize
\caption{\textbf{Ansatz: the twelve accounting exports.} Tokens in millions (W, workers; V, verifier; M, main agent; C, curator; four-role sum). Outcomes refer to full targets; Bernoulli's partial pass is included in the study total of ten but not the nine full solutions. Facts are cumulative project snapshots. Window, accounting minutes. USD, strategy-consultation charges only. $^{a}$Sum of the latest snapshot per project; continuation snapshots replace, rather than add to, the preceding snapshot for the same project.}
\setlength{\tabcolsep}{3.2pt}
\begin{tabular}{@{}lllrrrrrrrrrr@{}}
\toprule
Task & Phase & Outcome & Window & W & V & M & C & Sum & Rounds & Verifier runs & USD & Facts\\
\midrule
01 & first pass & solved & 49.2 & 51.67 & 0.97 & 0.35 & 0.08 & 53.07 & 618 & 29 & 0.55 & 28\\
02 & first pass & solved & 81.9 & 106.04 & 1.15 & 0.54 & 0.34 & 108.07 & 1,048 & 38 & 1.06 & 37\\
03 & first pass & not solved & 126.7 & 66.16 & 12.23 & 2.15 & 1.12 & 81.66 & 25 & 303 & 1.46 & 984\\
03 & resume (r2, live) & not solved & 1,608.9 & 89.36 & 77.30 & 58.69 & 15.04 & 240.39 & 109 & 1,725 & 3.09 & 4,102\\
04 & first pass & not solved & 356.2 & 157.19 & 23.06 & 2.40 & 0.25 & 182.91 & 23 & 514 & 2.51 & 467\\
04 & resume (r3) & solved & 122.2 & 83.76 & 12.00 & 1.07 & 1.27 & 98.10 & 469 & 288 & 3.40 & 586\\
05 & first pass & solved & 47.5 & 35.12 & 1.92 & 0.28 & 0.24 & 37.56 & 235 & 54 & 1.26 & 51\\
06 & first pass & solved & 27.9 & 10.66 & 0.80 & 0.52 & 0.18 & 12.16 & 98 & 29 & 1.05 & 29\\
07 & first pass & solved & 44.1 & 23.91 & 1.54 & 0.29 & 0.40 & 26.14 & 194 & 41 & 0.60 & 39\\
08 & first pass & solved & 44.0 & 22.43 & 2.01 & 0.25 & 0.28 & 24.98 & 97 & 52 & 0.83 & 47\\
09 & first pass & solved & 43.6 & 27.88 & 1.54 & 0.36 & 0.29 & 30.06 & 183 & 45 & 0.78 & 43\\
10 & first pass & solved & 44.4 & 24.00 & 2.03 & 0.58 & 0.30 & 26.91 & 104 & 55 & 1.80 & 51\\
\midrule
All twelve & & 9 solved & 2,596.6 & 698.18 & 136.55 & 67.47 & 19.80 & 922.00 & 3,203 & 3,173 & 18.39 & 5,013$^{a}$\\
First pass (10) & & 8 solved & 865.5 & 525.06 & 47.26 & 7.72 & 3.48 & 583.52 & 2,625 & 1,160 & 11.90 & \\
Resumes (2) & & 1 solved & 1,731.1 & 173.12 & 89.30 & 59.75 & 16.31 & 338.49 & 578 & 2,013 & 6.49 & \\
\bottomrule
\end{tabular}
\end{table}

\subsection{Complete ablation outcomes}
Supplementary Table~13 gives the complete ten-task outcomes displayed in main-text Fig.~\ref{fig:audit}b. The reference column gives the Ansatz result under the study criterion; A--D are the four ablations. Audit-flagged outcomes are excluded from retained totals.

\stab{13}
\begin{table}[ht]
\centering\small
\caption{\textbf{Complete-system reference and ablation outcomes.} S, solved; P, Bernoulli partial pass under the study criterion; --, not solved; F, initially marked solved but excluded after re-review. A, flat fact notebook; B, no exploration graph or curator; C, worker self-review; D, no strategy consultation. Ansatz settings are listed in Supplementary Table~11; ablation settings are described below.}
\begin{tabularx}{\linewidth}{@{}Xccccc@{}}
\toprule
Task & Ansatz & A & B & C & D\\
\midrule
01 Automorphism & S & S & S & S & S\\
02 M\"obius & S & S & S & S & S\\
03 Bernoulli & P & -- & -- & -- & --\\
04 Dilation & S & -- & -- & -- & --\\
05 SPDE & S & S & -- & F & S\\
06 Tree lattice & S & S & S & S & --\\
07 Matching & S & S & S & S & S\\
08 Dressian & S & S & S & S & S\\
09 Hilbert & S & S & S & S & S\\
10 von Neumann & S & S & S & -- & --\\
\midrule
Retained passes & 10/10 & 8/10 & 7/10 & 6/10 & 6/10\\
\bottomrule
\end{tabularx}
\end{table}

\subsection{Ablation configurations and assessment}
The four ablation arms used a common five-worker configuration on all ten tasks. A replaced content-addressed fact records and predecessor edges with a sequential notebook while retaining the verifier. B disabled exploration-graph tools and the curator. C had proposing workers apply the verifier's review prompts and submit their own verdicts, bypassing the separate verifier service and its prechecks while retaining structured writes and verdict traces. D disabled strategy consultation, leaving the main agent to provide guidance. The interventions and final outcomes correspond to main-text Fig.~\ref{fig:audit}b.

\textit{Configuration and assessment.} All ablations used five workers (\code{high:2,xhigh:3}) with GPT-5.6 Sol as the worker/verifier model. Web access and cross-project recall were off. Main agents and curators ran as Claude Code sessions, except that B had no curator. Final retained totals were 8, 7, 6 and 6 for A--D. For C, fresh-verifier review followed solving without repair feedback; the SPDE result was excluded, reducing seven initially solved targets to six retained targets. Some dependencies were sampled rather than reviewed exhaustively. The Ansatz reference uses a different worker allocation and stopping policy, so it is not a matched control for isolating individual component effects.

\subsection{Self-review audit}
Worker self-review rejected 3 of 831 submissions and admitted 828 records across the ten tasks. Three M\"obius admissions were revoked before re-review, leaving 21 available records for that task. A fresh verifier reviewed 206 selected records, accepting 186 and rejecting 20 (main-text Fig.~\ref{fig:audit}c; Supplementary Table~14). Selection covered all available records for tasks 01, 02, 06 and 07, and named targets plus 12 records selected by a fixed identifier stride for each remaining task. The reviewer did not return repair feedback to the original runs. The rejection fraction describes this selected audit, not a random-sample estimate. Final-target and dependency assessment flagged SPDE in arm C; retained task outcomes are shown separately in Supplementary Table~13. A rejected dependency indicates that the recorded argument requires reassessment, not that every downstream statement is false.

\stab{14}
\begin{table}[ht]
\centering\small
\caption{\textbf{Self-review audit.} Top, cumulative records admitted under self-review and their post hoc re-review by a fresh verifier across all ten tasks; three M\"obius admissions were revoked before the audit. Bottom, examples of rejected records (identifier, whether a named target, review time in seconds).}
\begin{tabularx}{\linewidth}{@{}L{2.8cm}rrrrrY@{}}
\toprule
Task & Admitted & Re-reviewed & Accepted & Rejected & Targets accepted & Selection\\
\midrule
01 Automorphism & 22 & 22 & 21 & 1 & 4/4 & census\\
02 M\"obius & 24 & 21 & 21 & 0 & 6/6 & census\\
03 Bernoulli & 103 & 22 & 21 & 1 & 10/10 & named targets + 12 by fixed ID stride\\
04 Dilation & 235 & 12 & 8 & 4 & -- & named targets + 12 by fixed ID stride\\
05 SPDE & 109 & 23 & 20 & 3 & 9/11 & named targets + 12 by fixed ID stride\\
06 Tree lattice & 15 & 15 & 15 & 0 & 4/4 & census\\
07 Matching & 23 & 23 & 20 & 3 & 14/14 & census\\
08 Dressian & 67 & 17 & 17 & 0 & 5/5 & named targets + 12 by fixed ID stride\\
09 Hilbert & 46 & 30 & 27 & 3 & 16/18 & named targets + 12 by fixed ID stride\\
10 von Neumann & 184 & 21 & 16 & 5 & 8/9 & named targets + 12 by fixed ID stride\\
\midrule
Total & 828 & 206 & 186 & 20 & 76/81 & \\
\bottomrule
\end{tabularx}
\vspace{2mm}
\begin{tabular}{@{}llll@{}}
\toprule
Task & Record & Named target & Seconds\\
\midrule
01 & \texttt{179c3b1daf91f661} & no & 97\\
03 & \texttt{b3af54f824e5017f} & no & 263\\
10 & \texttt{337678f5777a0098} & yes & 407\\
10 & \texttt{001de18dfc3f334f} & no & 112\\
10 & \texttt{775cfe649186300f} & no & 263\\
10 & \texttt{d260b81b0621cd11} & no & 160\\
10 & \texttt{ec86561438b78250} & no & 142\\
\bottomrule
\end{tabular}
\end{table}

\subsection{Ansatz retrieval activity}
Supplementary Table~15 gives the per-problem retrieval counts shown in main-text Fig.~\ref{fig:audit}f. Ansatz recorded 834 neighbourhood-retrieval calls and 315 cross-project-recall calls. SPDE accounted for the largest count in each channel, with 343 and 155 calls, respectively. Neighbourhood retrieval supplies context from the current project's proof dependencies and exploration relations, while cross-project recall provides statements, negative findings and methodological lessons from accessible projects. These counts measure memory access; establishing successful mathematical reuse requires tracing retrieved material into subsequent arguments.

\stab{15}
\begin{table}[ht]
\centering\small
\caption{\textbf{Ansatz retrieval activity on First Proof Second Batch.} Per-problem counts of neighbourhood-retrieval and cross-project-recall calls, corresponding to main-text Fig.~\ref{fig:audit}f.}
\begin{tabularx}{\linewidth}{@{}Xrr@{}}
\toprule
Task & Neighbourhood-retrieval calls & Cross-project-recall calls\\
\midrule
01 Automorphism & 20 & 5\\
02 M\"obius & 26 & 5\\
03 Bernoulli & 109 & 16\\
04 Dilation & 73 & 23\\
05 SPDE & 343 & 155\\
06 Tree lattice & 99 & 39\\
07 Matching & 57 & 41\\
08 Dressian & 53 & 6\\
09 Hilbert & 32 & 13\\
10 von Neumann & 22 & 12\\
\midrule
Total & 834 & 315\\
\bottomrule
\end{tabularx}
\end{table}
\FloatBarrier

\subsection{Third Batch: evaluation protocol and accounting}

We evaluated the harness on First Proof Third Batch, ten research-level problems whose solutions were released
on 14 October 2026 after an expert-refereed evaluation of five external systems~\cite{si-firstproof3}. We used Ansatz with the same offline protocol as in the Second Batch and the batch-specific settings described below. A second language model assessed the recorded results against the published answers in separate grading sessions; those grades are statement-level comparisons and not proof reviews.

\textit{Tasks and external results}

The ten local problem pages reproduce pages 13--23 of the First Proof Third Batch report, with the report's
pre-registered statements omitted from the input so that the per-problem sessions never saw the intended answer.
External results are the report's correctness scores (its Table~2), on a scale from $-1$ to $1$; the report
counts a submission as correct when it scores $0.5$ or better. The five systems were run by the organizers on
the ten-problem input under a 24-hour, one-shot limit with two or three referees per submission: GPT-6 Astra
scored 7 correct, ProofCouncil 9, MerLean 4, the Claude Opus 5.5 skill Siege 3 (it declined to submit
problems 3, 4, 8, 9 and 10, each scored zero), and the open-weight system Humanize 0. Three of the five external
systems (GPT-6 Astra, ProofCouncil, MerLean) ran with web search on.

\textit{Configurations}

All runs used the persistent harness in a single isolated deployment, in \textbf{benchmark mode} with the same
offline restrictions as Ansatz in the Second Batch. The Third Batch worker and verifier model was
\texttt{openai.gpt-6-astra}, with strategy consultation through the same model at maximum effort without tools, web
search and fetch denied to every role including the main agent and curator, and a sequencer launching a fresh
main-agent session per problem. The roster was
\texttt{high:3,xhigh:5,max:1} (nine seats), compared with ten seats for Ansatz in the Second Batch.
Initial runs were time-capped, with continuations retaining the same project state after the cap was lifted.

The two problems not solved within the cap (4 and 9) were \textbf{resumed with the cap lifted}: the same
projects, their verified facts and seats kept, a 24-hour backstop, and the early-stop rule switched off.
Problem~4's solution during continuation followed a within-run strategy consult that corrected the induction target; we do
not count that as an operator intervention, but the cap lift itself is one. 

\textit{Endpoints and accounting}

The harness solved nine full targets in the Third Batch; geodesic boundaries remained unsolved. The accounting-window and token conventions follow those used for the Second Batch. Three batch-specific accounting notes apply. One verifier run lacks a token line (problem~4, capped
pass), so that total is recorded rather than complete. The resumes of problems 4 and 9 reused the capped pass's
round-log filenames, so for those two problems the worker tokens of the resume cannot be separated from the
capped pass; their capped-pass per-seat accounting survives only in the exports taken before the resumes, and
only the verifier, main-agent and curator additions of the resumes can be reported separately. Consultation
ledgers are project-cumulative, so the resumed-run exports for problems 4 and 9 repeat their capped-pass
consults; the reported Third Batch totals count each consult once.

Grading uses separate read-only sessions that compare the run's recorded statements
with the problem's pre-registered statement and the referee summaries, and record whether the conclusions
agree, whether that agreement rests on cited inputs the run could not check, whether the run proved strictly
less, or whether the two contradict each other. Its inputs contain the answers and are forbidden to the
per-problem sessions, which never saw them. An audit of the run records confirms the offline protocol held: zero
worker web searches, zero verifier web-search records, and zero consult tool calls.

\section{Supplementary Note 8. Software verification and capture provenance}
\stab{16}
\begin{table}[ht]
\centering\small
\caption{\textbf{Regression results at the pinned snapshot} (commit \code{b30740e}, 13 October 2026). Subtests are reported separately from test cases. No paid model run was initiated.}
\begin{tabularx}{\linewidth}{@{}Yrrr@{}}
\toprule
Suite & Tests passed & Subtests passed & Skipped\\
\midrule
Console and application (Python) & 2,457 & 1,092 & 8\\
Standalone CGM (Python) & 535 & -- & 0\\
JavaScript contracts & 95 & -- & 0\\
\bottomrule
\end{tabularx}
\end{table}
Console tests use temporary data roots and synthetic dependencies; the standalone suite uses a separate environment; browser checks exercise JavaScript contracts without model requests. Tested families include parsing, admission, budgets, concurrency, artefact correspondence, manuscript ingestion, verdict handling, route policy and interpretation of formal results. Of the eight skips, three require checker sockets, two a pinned external archive, one an optional engine environment, one a production checkout and service-account layout, and one the absence of a parser package that was in fact installed, so the suites do not retest the live broker or every optional integration. Separately, the supplied harness snapshot passed its 52 existing offline tests (15 core-memory, 5 neighbourhood, 5 exploration-graph and 27 precheck tests) without model calls; these check mechanics, not research results, and it is not established that this snapshot is the exact code that ran the benchmark.

\textit{Lean recheck.} We copied the exact exported Lean source of the public odd-integer artefact, checked its SHA-256 hash and ran Lean 4.14.0 (commit \code{410fab728470}) with Mathlib \code{4bbdccd9c5f8} directly, without a provider or production-broker request. The probe checked the type of \code{main} and the axiom dependencies of all four declarations; it exited with status zero and no error output. Three declarations, including the target, used \code{propext}; integer closure used no axioms. The run took 50.107\,s including the probe. The result applies to that artefact and installation; it does not validate later revisions or the deployed checking boundary.

\textit{Captures.} The capture environment serves the original console against isolated state and a synthetic account with mutation routes disabled; predetermined model replies exercise the ordinary CGM controller, predetermined reviews annotate the parity manuscript, and a display-only replay projects the published formal source and its dated checking metadata into the interface. Public catalogue and atlas captures use a local static build of the pinned frontend and its bundled datasets with external requests and live overlays disabled. Captures use a 1,500$\times$1,000 CSS-pixel viewport at device scale factor 1.5 (2,250$\times$1,500 pixels), and a manifest records the selected control, timestamp, source identity and screenshot hash. User-selected snapshots are replayed through the revised read-only viewer at commit \code{c37224c} using loopback requests only. Figure crops change framing, not pixels.

\section{Supplementary Note 9. Open problems: statements and claimed results}
\label{sec:si-open-problems}
Each paragraph describes a candidate result reported by the corresponding harness project in main-text Table~1. Full and Partial refer to the scope of the reported result; Lean and AI-proposed describe its recorded checking status. The source reports Lean formalizations for Jamison and problems 289, 488, 348 and 1063, including the partial result for 1063. These research formalizations were not independently rebuilt for this paper. The remaining results were accepted by the harness's model-based review. Problem numbers refer to the Erd\H{o}s problems catalogue~\cite{si-bloomerdos}.

\textit{Jamison caterpillar conjecture.} The mean subtree order of a tree is the average order of its nonempty connected induced subgraphs. Jamison conjectured that every maximizer among trees with $n$ vertices is a caterpillar~\cite{si-Jamison83,si-Jamison84}; Mol and Oellermann verified this for $n\le24$ and restricted the limbs of any maximizer~\cite{si-MolOellermann19}. The manuscript reports a proof that every $n$-vertex maximizer is a caterpillar: a shortest terminal arm is reduced to a broom, which is then straightened by exact polynomial certificates.

\textit{Problem 561.} For star forests $F_1=\bigcup_i K_{1,n_i}$ and $F_2=\bigcup_j K_{1,m_j}$, let $L$ be the sum of the diagonal maxima $l_k=\max\{n_i+m_j-1:i+j=k\}$. Burr, Erd\H{o}s, Faudree, Rousseau and Schelp conjectured that the size Ramsey number satisfies $\hat R(F_1,F_2)=L$~\cite{si-BEFRS78}; the formula was known for uniform forests, for a single star, for two equal stars and when every size is odd~\cite{si-DJKR25}. The manuscript reports a full proof of the conjectured formula for every pair of star forests. The result is classified as Full and AI-proposed: it has passed the harness's model-based review and has not been formalized in Lean.

\textit{Problem 289.} Erd\H{o}s and Graham asked whether 1 can be written as a sum of reciprocals over $k$ finite intervals of positive integers for every sufficiently large $k$~\cite{si-ErGr80}. The catalogue form requires the intervals to be pairwise disjoint and non-adjacent, each of length at least two. The manuscript reports a proof that such a threshold $k_0$ exists, using a Pell anchor, equal-mass packets and a summable reserve; the threshold is not effective, because the last step uses compactness. The unrestricted variant, which allows overlap, already had a short affirmative argument~\cite{si-Kovac289}, and a public manuscript claims a stronger version in which every interval contains two or three integers~\cite{si-Tang289}.

\textit{Problem 488.} For a finite nonempty set $A$ of positive integers, let $B$ be the set of their multiples and $d_A(t)=|B\cap[1,t]|/t$. Erd\H{o}s asked whether $d_A(m)<2d_A(n)$ whenever $m>n\ge\max A$. The manuscript reports a finite construction of $A$ and admissible $m,n$ with $d_A(m)/d_A(n)>256/67$. Its elements have prime factors in a fixed finite set and are at least a threshold, but division by any of their prime factors puts them below it; their multiples therefore have low density near $\max A$, while an exact residue-class count gives a higher density at a later cutoff.

\textit{Problem 348.} A sequence is complete if every sufficiently large positive integer is a sum of terms at distinct indices; repeated values are allowed. The question asks which $0\le m<n$ admit a sequence that stays complete after every deletion of $m$ terms but becomes incomplete after every deletion of $n$ terms~\cite{si-ErGr80}. The cases $m=0,1$ were known, and van Doorn excluded $m\ge2$ under the stronger definition that requires every positive integer to be represented. The manuscript reports a proof that, under eventual completeness, the admissible pairs are exactly $(0,n)$ for $n\ge1$ and $(1,n)$ for $n\ge2$: powers of two and the Fibonacci sequence supply the two families, and any sequence that survives every two-term deletion admits a completeness-preserving deletion of any prescribed finite size.

\textit{Problem 811.} An edge colouring of $K_n$ with $m=e(G)$ colours is balanced when $n\equiv1\pmod m$ and every vertex meets every colour equally often; Erd\H{o}s, Pyber and Tuza asked which graphs $G$ appear as rainbow subgraphs in every large balanced colouring. Axenovich and Clemen showed that most cliques fail and conjectured that $K_q$ fails for every $q\ge4$, and Clemen and Wagner settled $q=4$ (refs~\citenum{si-AxCl24,si-ClWa23}). The manuscript gives, for every $q\ge4$, a finite balanced base colouring with no rainbow $K_q$ and lifts it to arbitrarily large admissible orders by the amplification of Axenovich and Clemen; for $q\ge12$ the base pairs cyclic distances on $\mathbb{Z}_N$ with $N=4\binom{q}{2}+1$, and the eight bases with $q\le11$ were checked by exhaustive search. The question for general graphs, including $C_6$, remains open.

\textit{Problem 644.} Let $f(k,7)$ be the largest transversal number of a $k$-uniform family in which every seven members have a two-point transversal; the conjectured order is $3k/4$. Fon-Der-Flaass, Kostochka and Woodall proved $f(k,7)\le\lceil7k/8\rceil$~\cite{si-FDFKW00}. For $k\ge150$ the manuscript reports $f(k,7)\le\lceil(6k+3)/7\rceil+1$, hence $\limsup_{k} f(k,7)/k\le6/7$.

\textit{Problem 993.} Erd\H{o}s asked whether the independence sequence $i_k(T)$ of every tree, or forest, is unimodal. Basit and Galvin showed that it is nondecreasing through roughly $n/8$ for trees~\cite{si-BaGa21}. The manuscript reports a proof that the sequence of every forest increases strictly through the first $\tfrac{131}{520}n$ terms and, for forests of maximum degree at most four, through $0.276n$, which is sharp because the path stops exactly there. The conjecture itself remains open.

\textit{Problem 1063.} Let $n_k$ be the least $n\ge2k$ for which exactly $k-1$ of $n,n-1,\dots,n-k+1$ divide $\binom{n}{k}$. Monier proved $n_k\le k!$ for $k\ge3$, Cambie improved this to $n_k\le\exp((1+o(1))k)$ (as recorded in the catalogue), and an AI-generated report claims $n_k\le\exp(Ck\log\log k/\log k)$ for an unspecified constant $C$~\cite{si-EPAD1063}. For sufficiently large $k$ the manuscript reports
\begin{equation*}
\exp\bigl(10^{-30}(\log k)^2\bigr)<n_k\le\exp\Bigl(\frac{k}{\log k}\bigl[\log\log k+\log\log\log k+\log2-\eta+o(1)\bigr]\Bigr)
\end{equation*}
for an absolute constant $\eta>0$, which gives a super-polynomial lower bound and refines the claimed upper bound; a matching growth estimate remains open.

\textit{Problem 302.} Let $f(N)$ be the largest size of a subset of $\{1,\dots,N\}$ that contains no distinct $a,b,c$ with $1/a=1/b+1/c$~\cite{si-ErGr80}. The catalogue records Cambie's asymptotic lower coefficient $5/8$ and van Doorn's upper coefficient $9/10$, and Khanukov's preprint gives a non-explicit gain over $5/8$ and an upper coefficient below $0.860853$~\cite{si-Khanukov26}. The manuscript reports $f(N)=cN+o(N)$, a characterization of $c$ as an infimum of approximations over finite sets of primes with an explicit convergence rate, and the bound $c\ge5/8+\exp(-100000000030)$. The exact value of $c$ remains open.

\textit{Conway's thrackle conjecture.} A thrackle is a drawing of a graph in the plane in which every two edges meet exactly once, adjacent edges at their common endpoint and independent edges at a proper crossing; Conway conjectured that every thrackle has at most as many edges as vertices. Lov\'asz, Pach and Szegedy proved $|E|\le2|V|-3$ (ref.~\citenum{si-LPS97}), later work brought the bound to $|E|\le1.393(|V|-1)$ (ref.~\citenum{si-Xu21}), and a preprint of 13 October 2026 claims $|E|\le1.385(|V|-1)$ for $|V|\ge4$ (ref.~\citenum{si-Scoppetto26}). The project, still running on 22 October 2026 and without a manuscript yet, released a verifier-accepted theorem that every thrackle satisfies $|E|\le29|V|/22$, and $22|E|\le29(|V|-1)$ unless the graph is a triangle or a triangle with one pendant edge. The conjecture itself remains open.

\textit{Graceful tree conjecture.} A graceful labelling of a tree with $m$ edges is a bijection from its vertices onto $\{0,1,\dots,m\}$ under which the absolute differences across the $m$ edges are exactly $1,\dots,m$; the graceful tree conjecture of Ringel, Kotzig and Rosa states that every finite tree has one~\cite{si-Ringel64,si-Rosa67}. Huang, Kotzig and Rosa proved it for trees with at most four leaves~\cite{si-HKR82}, Hrn\v{c}iar and Haviar for trees of diameter at most five~\cite{si-HrHa01}, and Panpa and Poomsa-ard for spiders, trees with at most one vertex of degree three or more, with at most four legs of length at least two~\cite{si-PaPo16}; exhaustive search covers all trees with at most 35 vertices~\cite{si-Fang10}. The project, which has no manuscript yet, holds verifier-accepted proofs that every tree with at most seven leaves is graceful (\code{a7c43c6607f096f8}) and that every spider with at most eight legs of length at least two, and any number of legs of length one, is graceful (\code{5a215aff7485c9d4}); the source records dependency closures of 1,587 and 1,609 facts, respectively. No proof has been checked by a person, and the conjecture itself remains open.

\section{Supplementary Data and Software}
The accompanying files are supplied in the arXiv ancillary directory \nolinkurl{anc/}. Its \nolinkurl{README.txt} describes archive coverage, retained filenames and omissions. The ancillary archives contain the archived non-TeX code, inputs and associated assets; separate TeX figure sources are retained in the companion materials rather than embedded in these ancillary archives.

\textbf{Supplementary Data 1} (\nolinkurl{anc/SupplementaryData1_records.zip}). Normalized records (\nolinkurl{reported_results.json}, \nolinkurl{fpbmain_results.json}, \nolinkurl{ansatz_table_results.json} and \nolinkurl{firstproof_second_batch_baselines.json}), the archived per-run accounting exports and post-hoc judging files, and the scripts that rebuild the normalized records without model calls.

\textbf{Supplementary Data 2} (\nolinkurl{anc/SupplementaryData2_run_reports.zip}). Operator reports used for these analyses: the Ansatz run ledger, summary and memory-activity records, and the ablation-study summaries (some in Chinese, as recorded).

\textbf{Supplementary Data 3} (\nolinkurl{anc/SupplementaryData3_prompt_protocol.pdf}). The 47 source records of the shared benchmark prompt protocol (main-agent entry contract, four role contracts, 17 worker skills, 3 verifier skills, runbook, ten kickoff prompts, ten problem statements and the verifier wrapper), reproduced as archived; they are not transcripts of any run and are not established to be the exact prompt revision of the Ansatz deployment.

\textbf{Supplementary Software} (\nolinkurl{anc/SupplementarySoftware_figures.zip}). Archived non-TeX code and inputs for the original data displays. Current Python generation files and associated figure assets are supplied in \nolinkurl{anc/Current_figure_sources.zip}; the available spreadsheets and presentation CSV inputs are supplied in \nolinkurl{anc/Source_Data.zip}. Archived data files retain their original field names and identifiers. The ancillary README documents the available coverage and the separate companion materials.

\begingroup
\footnotesize
\setlength{\bibsep}{1pt}
\renewcommand{\refname}{Supplementary References}

\endgroup

\fi
\end{document}